\documentclass[11pt]{article}
\usepackage[utf8]{inputenc}
\usepackage[margin=1in]{geometry}
\usepackage{graphicx}
\usepackage{booktabs}
\usepackage{longtable}
\usepackage{array}
\usepackage{multirow}
\usepackage{amsmath}
\usepackage{amsfonts}
\usepackage{url}
\usepackage[super,comma,sort&compress]{natbib}
\usepackage{hyperref}
\usepackage{pdflscape}
\usepackage{afterpage}
\usepackage{float}
\usepackage[table]{xcolor}
\usepackage{tikz}
\usetikzlibrary{positioning, arrows, shapes}

\title{The Validity Gap in Health AI Evaluation: A Cross-Sectional Analysis of Benchmark Composition}

\author{Alvin Rajkomar, M.D.$^{1,*}$, Pavan Sudarshan, B.S.$^{1}$, Angela Lai, M.S.E.$^{1}$, Lily Peng, M.D.,Ph.D.$^{1}$}

\date{\today}

\begin{document}

\maketitle

% Manual footnotes without automatic symbols
\renewcommand{\thefootnote}{\fnsymbol{footnote}}
\footnotetext[1]{$^{1}$Apple}
\footnotetext[0]{$^{*}$Corresponding author: Alvin Rajkomar can be contacted at alvinr@apple.com}
\renewcommand{\thefootnote}{\arabic{footnote}}

\begin{abstract}
% Abstract — unstructured, argument-forward

Benchmarks for consumer-facing health large language models (LLMs) report aggregate performance scores but rarely characterize the queries they contain---analogous to publishing clinical trial results without describing inclusion criteria. This gap is especially consequential because consumers, unlike clinicians, lack the domain expertise to critically evaluate AI-generated health information. We applied a standardized 16-field query profiling taxonomy to 18,707 consumer health queries across six public benchmarks and identified four systematic blind spots that undermine evaluation validity. First, benchmarks defaulted to a ``standard adult'' profile, with pediatric and older-adult queries comprising less than 11\% of the corpus despite these groups accounting for over half of clinical encounters, while globally significant conditions such as malaria and tuberculosis were effectively absent. Second, chronic disease management (5.5\%) and preventive care (2.7\%) were profoundly underrepresented relative to real-world primary care, where chronic conditions account for 39\% of office-based physician visits. Third, clinical document interpretation---the parsing of laboratory results, imaging reports, and medical record excerpts that patients increasingly bring to AI systems---was scarcely tested, with lab results in only 5.2\% of queries and raw clinical artifacts in 0.6\%. Fourth, behavioral health crisis scenarios---self-harm and suicidal ideation---comprised less than 0.7\% of all queries and were virtually absent from the conversational benchmarks where chatbots actually operate, leaving crisis de-escalation untested. These compositional gaps mean that benchmark performance may not generalize to the clinical contexts where AI systems are increasingly deployed. We propose routine query profiling---analogous to CONSORT reporting for clinical trials---to align health AI evaluation with the complexity of clinical practice, and release open-source tools to facilitate adoption.

\end{abstract}

\section{Introduction}
% Introduction — CONSORT analogy first, tightened

Clinical trials report participant characteristics. Consumer health AI benchmarks do not. This asymmetry is consequential: just as the generalizability of a trial depends on who was enrolled~\cite{rothwellExternalValidityRandomised2005}, the validity of extrapolating benchmark performance to clinical deployment depends on the composition of the queries used for evaluation. Without compositional transparency, aggregate accuracy metrics may conceal systematic gaps in the scenarios that matter most for patient safety---a problem analogous to publishing trial results without inclusion criteria, which CONSORT guidelines were designed to prevent~\cite{hopewellCONSORT2025Statement2025}.

The consumer context warrants particular scrutiny. When clinicians use AI-generated health information, they bring domain expertise that serves as a critical filter---they can recognize implausible outputs, weigh suggestions against clinical training, and contextualize recommendations within a patient's broader care. Consumers lack this interpretive safety net: they cannot readily distinguish a confidently stated but incorrect response from a sound one, making them disproportionately vulnerable to harm. Because consumer queries also differ in character---skewing toward symptom lookup and self-management rather than differential diagnosis or treatment optimization~\cite{costa-gomesHowPeopleUse}---benchmarks must be evaluated against the specific scenarios consumers actually encounter. The scale of this exposure is large and growing: approximately one-third of U.S.\ adults now report using AI tools for health information or advice, with disproportionate uptake among younger adults, uninsured individuals, and racial and ethnic minorities---often driven by difficulty accessing or affording traditional care~\cite{kffjulianmKFFTrackingPoll2026}. An analysis of over 500,000 health-related conversations on Microsoft Copilot found that nearly one in five involved personal symptom assessment or condition management, with usage peaking in evening hours when traditional care is least accessible~\cite{costa-gomesHowPeopleUse}. Benchmark composition profoundly affects model evaluation because different query types stress different capabilities~\cite{bediTestingEvaluationHealth2025, yanLargeLanguageModel2024, beanMeasuringWhatMatters2025}. A model achieving 85\% accuracy on a benchmark dominated by simple symptom lookups may perform far worse on complex clinical reasoning tasks requiring nuanced risk communication or integration of personal health context. The need for benchmark transparency aligns with emerging dataset documentation frameworks such as Data Nutrition Labels~\cite{hollandDatasetNutritionLabel2018} and Dataset Statements~\cite{benderDataStatementsNatural2018}, which emphasize reporting data distributions and intended use cases.

Health queries span a wide range of tasks---from diagnosis and treatment planning to fitness tracking, wellness advice, preventive care, and chronic disease management. Even within a single domain, structural complexity varies substantially: a concise question such as ``Do I have diabetes?'' poses a fundamentally different evaluation challenge than a detailed symptom timeline accompanied by laboratory values and imaging reports. Differences in information-seeking goals---symptom explanation, treatment guidance, triage disposition---further stratify model requirements.

We introduce query profiling: a standardized framework for characterizing consumer health queries across three dimensions---context (structural properties and information richness), topic (clinical domain and conditions), and intent (information-seeking goals). Applying this 16-field taxonomy to six widely used benchmarks spanning 18,707 consumer queries, we identify four systematic blind spots---in demographic representation, chronic disease management, clinical document interpretation, and behavioral health crisis scenarios---that collectively undermine the validity of current health AI evaluation. We release open-source tagging tools and tagged datasets enabling reproducible query profiling across research groups.

\section{Methods}
\subsection{Study Design and Taxonomy Development}
We performed a cross-sectional analysis of publicly available health query datasets to characterize the structural and clinical composition of benchmarks used to evaluate large language models (LLMs). To standardize this characterization, we developed a ``Query Profile,'' a 16-field taxonomy designed to capture three dimensions of health information-seeking behavior:
\begin{itemize}
    \item \textbf{Context:} Structural properties, including conversation depth (single query versus back-and-forth conversation), population characteristics, presence of objective data (e.g., laboratory values, vitals, clinical artifacts), and information richness.
    \item \textbf{Topic:} The primary clinical domain and specific medical conditions referenced.
    \item \textbf{Intent:} The user's information-seeking goal, classified into nine categories ranging from factual education to triage and chronic disease management.
\end{itemize}
In the absence of a standard ontology for characterizing health LLM queries, we synthesized a pragmatic taxonomy informed by clinical experience and commercial health information-serving ontologies. Traditional clinical ontologies such as ICD or SNOMED were not suitable because they presuppose confirmed diagnoses and clinical context that consumer queries typically lack; our taxonomy instead captures the limited, often ambiguous information available from a non-expert's perspective. This framework represents one reasonable approach; alternative categorizations may be appropriate for specialized clinical domains. The full taxonomy and definitions are provided in Table 1.

\begin{table}[H]
  \centering
  \caption{Query Profile Taxonomy: Elements and Descriptions}
  \label{tab:query_profile}
  \begin{tabular}{p{4cm}p{9cm}}
  \toprule
  \textbf{Element} & \textbf{Description} \\
  \midrule
  \multicolumn{2}{l}{\textbf{Context Dimensions}} \\
  Population & Age-specific demographic (pediatric, adult, older adult), derived from user-provided context cues. \\
  Conversation Type & Single question versus multi-turn conversation. \\
  Length Detail & Short text string versus extended narrative. \\
  Objective Data & Presence of quantitative health data (labs, vitals, imaging). \\
  Context Depth & Information richness (age, timeline, clinical anchors). \\
  Terminology Level & Lay versus technical terminology. \\
  Clarification Needed & Self-contained versus ambiguous query (e.g., ``ibuprofen dose for adults'' vs.~``ibuprofen info''). \\
  Setting & Implied clinical setting (outpatient, inpatient, emergency). \\
  User Type & Consumer versus healthcare professional. \\
  Language & English versus non-English. \\
  Region & Geographic or regulatory context. \\
  \midrule
  \multicolumn{2}{l}{\textbf{Topic Dimensions}} \\
  Topic Area Path & Hierarchical classification by body system or condition. \\
  Key Conditions & High-priority health conditions as defined by global health agencies. \\
  Specialty & Clinical specialty best aligned with the query. \\
  \midrule
  \multicolumn{2}{l}{\textbf{Intent Dimensions}} \\
  Intent (Top-level) & Primary information-seeking goal (9 categories). \\
  Intent (Sub-level) & Subcategory of intent (e.g., treatment, prognosis). \\
  Risk Sensitivity & Urgency and severity implied by the query. \\
  \bottomrule
  \end{tabular}
\end{table}

\subsection{Data Sources}
We analyzed six widely cited public benchmarks containing consumer-facing health queries ($N=20,034$). These datasets were selected to represent the diverse sources of health inquiries currently used to train or evaluate AI models:
\begin{itemize}
    \item \textbf{Search Engine Queries:} \textit{HealthSearchQA}~\cite{singhalLargeLanguageModels2023a} ($N=3,173$) and \textit{MashQA Test} ($N=3,491$), comprising short, atomic questions typical of web-based search.
    \item \textbf{Online Medical Forums:} \textit{MedRedQA Test}~\cite{nguyenMedRedQAMedicalConsumer2023} ($N=5,099$), consisting of user-authored narratives posted to physician-facing public forums, often containing detailed medical history.
    \item \textbf{Simulated Interactive Dialogue:} \textit{HealthBench Main}~\cite{aroraHealthBenchEvaluatingLarge2025} ($N=5,000$), representing multi-turn interactions between simulated users and AI agents.
    \item \textbf{Wearable Data Streams:} \textit{GoogleFitbit Sleep} ($N=1,521$) and \textit{GoogleFitbit Fitness} ($N=1,750$)~\cite{khasentinoPersonalHealthLarge2025}, representing inquiries derived from continuous biometric data streams rather than explicit text prompts.
\end{itemize}
For analysis, we grouped these benchmarks into three generations reflecting the evolution of health information-seeking: Generation~1 (search-engine queries: HealthSearchQA, MashQA), Generation~2 (forum-based case presentations: MedRedQA), and Generation~3 (interactive dialogue and wearable data: HealthBench Main, GoogleFitbit). This grouping facilitates comparison of clinical content across benchmark modalities.

All datasets were de-identified and publicly available; the study was exempt from institutional review board approval.

\subsection{Classification and Validation}
Each query was classified across the 16 taxonomic dimensions by GPT-5.2 (OpenAI), using strict definitions, controlled vocabularies, and deterministic decision rules (full prompt in Supplementary Material). This approach follows prior work using LLMs as standardized coding instruments for large-scale query analysis~\cite{chatterjiHowPeopleUse2025,costa-gomesHowPeopleUse}. For wearable-data datasets (GoogleFitbit), a rule-based programmatic classifier was used.

We assessed tagging validity through two complementary analyses. First, to establish reliability, we compared GPT-5.2 with Claude Opus~4.5 (Anthropic)---architecturally distinct models from independent organizations---on the HealthBench Main dataset ($N=4{,}967$ matched queries). Overall agreement averaged 90.0\% (Cohen's $\kappa=0.77$), with no significant distributional differences on any dimension (details in Supplementary Material). Second, to establish accuracy, two physicians independently reviewed a stratified sample of 80 queries, blinded to model identity, judging each tag's reasonableness on a 5-point Likert scale. Because this taxonomy is novel and the classification task has no pre-existing gold standard, clinician-judged reasonableness---whether a physician would consider the label appropriate---is the most relevant validation criterion. Clinicians rated tags as reasonable (Likert $\geq 4$) in 97\% of cases across all dimensions (mean 4.84--4.90/5), including on queries deliberately selected for model disagreement. Neither model was preferred on disagreement cases (sign test $p > 0.35$; Supplementary Material).

\subsection{Flow of Data}

\begin{figure}[H]
\centering
\includegraphics[width=1.0\textwidth]{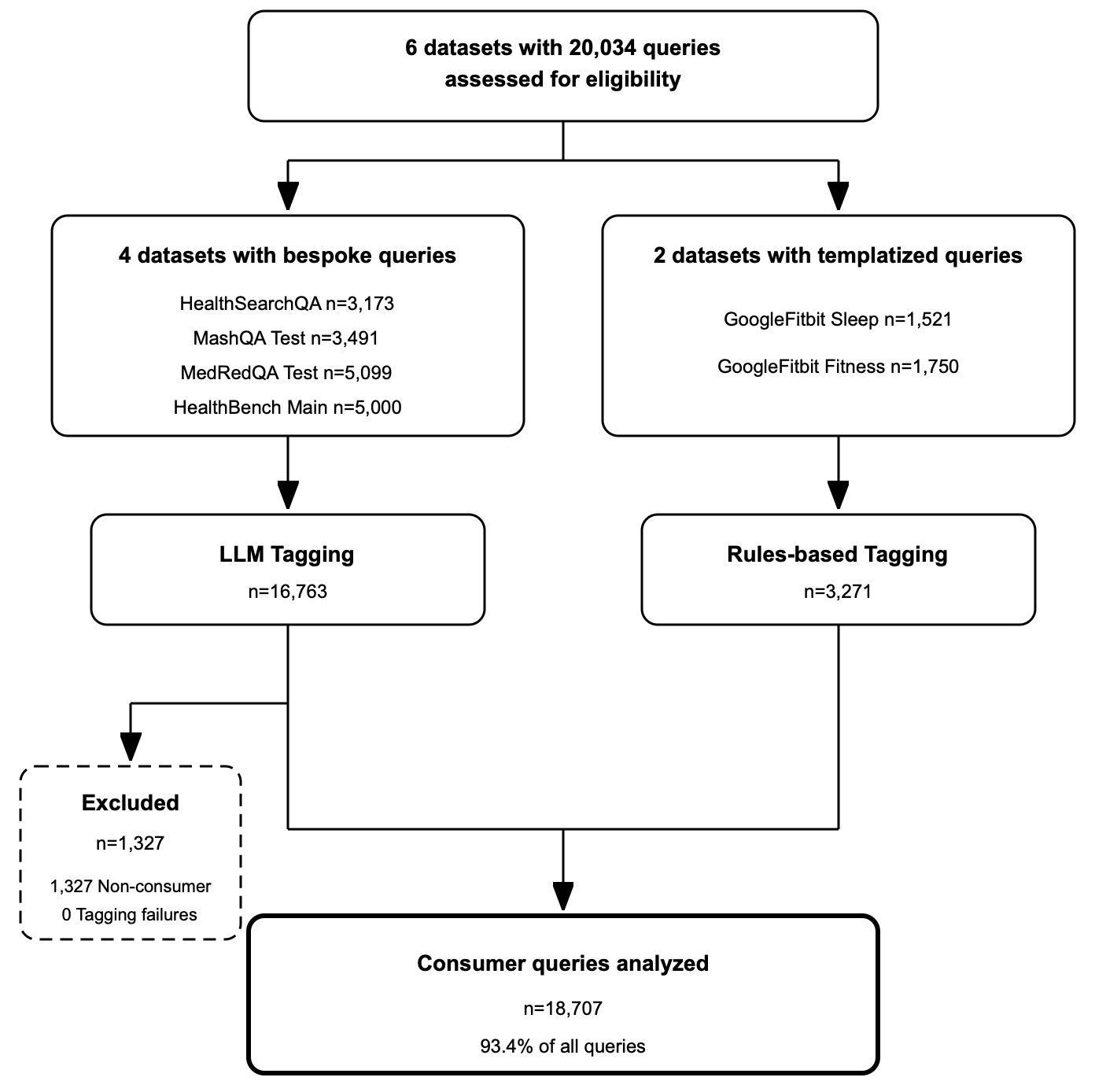}
\caption{CONSORT-style flow diagram showing study progression from initial dataset assessment through tagging methods to final analytic cohort of 18,707 consumer health queries.}
\label{fig:consort}
\end{figure}

A CONSORT-style diagram (Figure~\ref{fig:consort}) summarizes dataset selection, exclusions, and the final analytic sample.

\subsection{Statistical Analysis}
Descriptive statistics were used to characterize the distribution of context, topic, and intent across the datasets. We analyzed the prevalence of specific clinical needs---such as preventive care, chronic disease management, and high-acuity triage---to assess the alignment between benchmark composition and real-world clinical complexity. We compared benchmark distributions against national ambulatory care data~\cite{ashmanProductsDataBriefs2021, jettyEvaluationDeclinePrimary2025}. This reporting follows STROBE guidelines for observational research. All analyses used publicly available, de-identified datasets; no human subjects were involved.

\section{Results}
% Results — blind spots as primary findings

\subsection{Study Population}

Across the six benchmarks, we analyzed 20,034 total queries. After excluding 1,327 queries classified as originating from healthcare professionals, the final analytic cohort consisted of 18,707 consumer-facing health queries (93.4\% of the initial sample). The population was predominantly adult (89.0\%), English-speaking (96.4\%), and characterized by lay language (93.2\%). Most queries (72.7\%) were classified as low-risk, and 91.9\% were single-turn interactions. Applying the query profiling taxonomy revealed four systematic blind spots in benchmark composition.

\subsection{Blind Spot 1: Demographic and Geographic Skew}

\begin{figure}[H]
\centering
\includegraphics[width=1.0\textwidth]{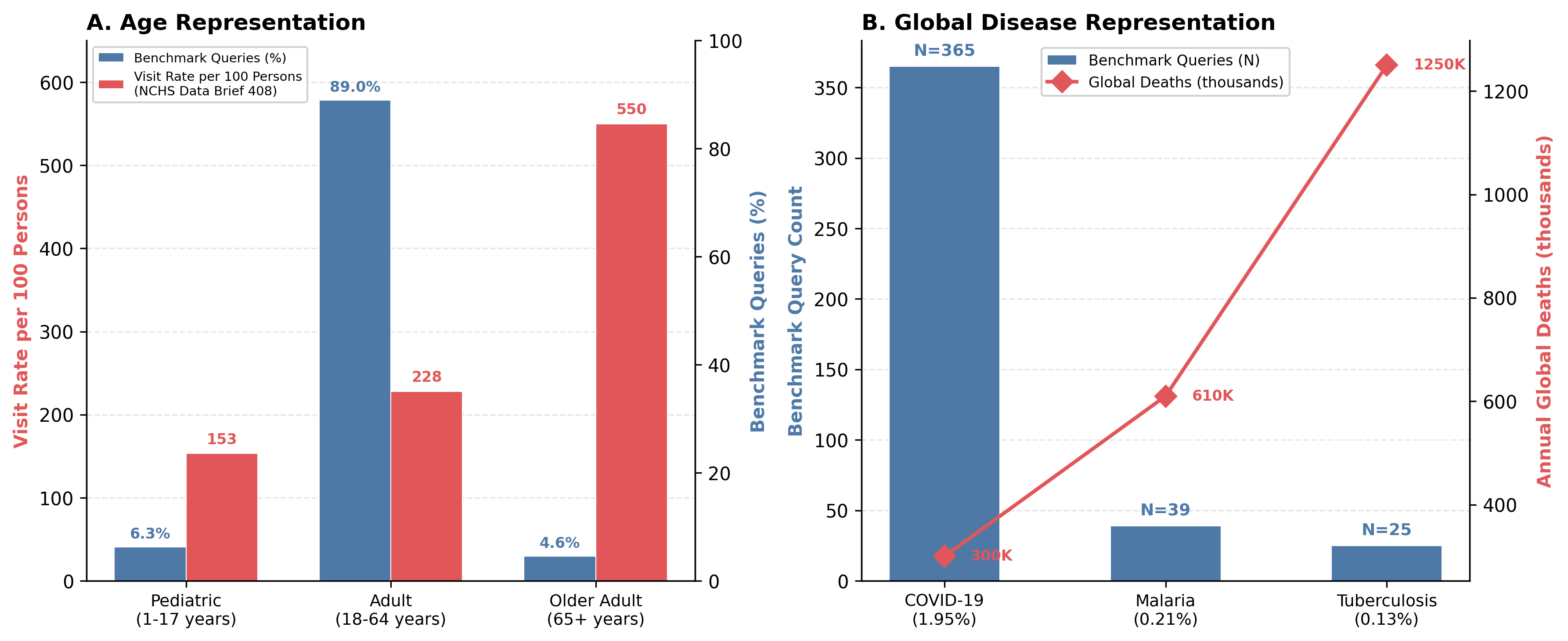}
\caption{Demographic and geographic representation gaps. (A)~Age distribution: benchmarks default to a ``standard adult'' (89\%), while older adults ($>$65 years)---who have the highest office-based physician visit rate (550 per 100 persons)---represent only 4.6\% of benchmark queries, and pediatric populations are similarly underrepresented. Visit rates from NCHS Data Brief No.\ 408 (2021). The visit rate for adults aged 18--64 (228 per 100 persons) is a derived aggregate of the 18--44 and 45--64 age brackets provided in the original NCHS brief. (B)~Global disease representation: despite causing over 1.8 million deaths annually (WHO Global TB Report 2024; WHO World Malaria Report 2024), malaria ($N=39$) and tuberculosis ($N=25$) are nearly absent from benchmarks, while COVID-19 ($N=365$) receives disproportionate coverage relative to its current global mortality burden.}
\label{fig:demographic_skew}
\end{figure}

Current benchmarks validate performance for a ``standard adult'' that does not reflect the populations most reliant on healthcare (Figure~\ref{fig:demographic_skew}). Queries explicitly referencing \textbf{older adults} ($>$65 years) comprised only 4.6\% ($N=861$) of the corpus, while \textbf{pediatric} queries (all age groups combined) represented 6.4\% ($N=1,190$). Together, these vulnerable populations accounted for less than 11\% of benchmark queries.

These proportions contrast starkly with healthcare utilization patterns. In the United States, adults aged 65 and over have the highest office-based physician visit rate (550 per 100 persons), more than three times the rate for adults aged 18--44 (173 per 100 persons)~\cite{ashmanProductsDataBriefs2021}. The remaining 89\% of benchmark queries assume a generic adult.

The geographic skew was equally pronounced. While \textbf{COVID-19} was relatively well-represented ($N=365$; 2.0\%), globally significant infectious diseases were effectively absent: \textbf{malaria} ($N=39$; 0.21\%) and \textbf{tuberculosis} ($N=25$; 0.13\%)---conditions responsible for over 1.8 million deaths annually, predominantly in low- and middle-income countries---appeared in negligible numbers. Current benchmarks thus primarily validate model safety for a narrow demographic: working-age adults in high-income, English-speaking regions.

\subsection{Blind Spot 2: Chronic Care and Prevention Underrepresentation}

Chronic disease management accounted for only 5.5\% ($N=1,037$) of all query intents, and preventive care constituted just 2.7\% ($N=513$). Within chronic care, the specific conditions that dominate primary care were strikingly rare: \textbf{diabetes} ($N=244$; 1.3\%), \textbf{hypertension} ($N=139$; 0.7\%), and \textbf{obesity} ($N=40$; 0.2\%)---combined, less than 3\% of the corpus. In contrast, acute symptom checking ($N=4,523$; 24.2\%) and health education ($N=5,473$; 29.3\%) together accounted for more than half of all queries.

\begin{figure}[H]
\centering
\includegraphics[width=1.0\textwidth]{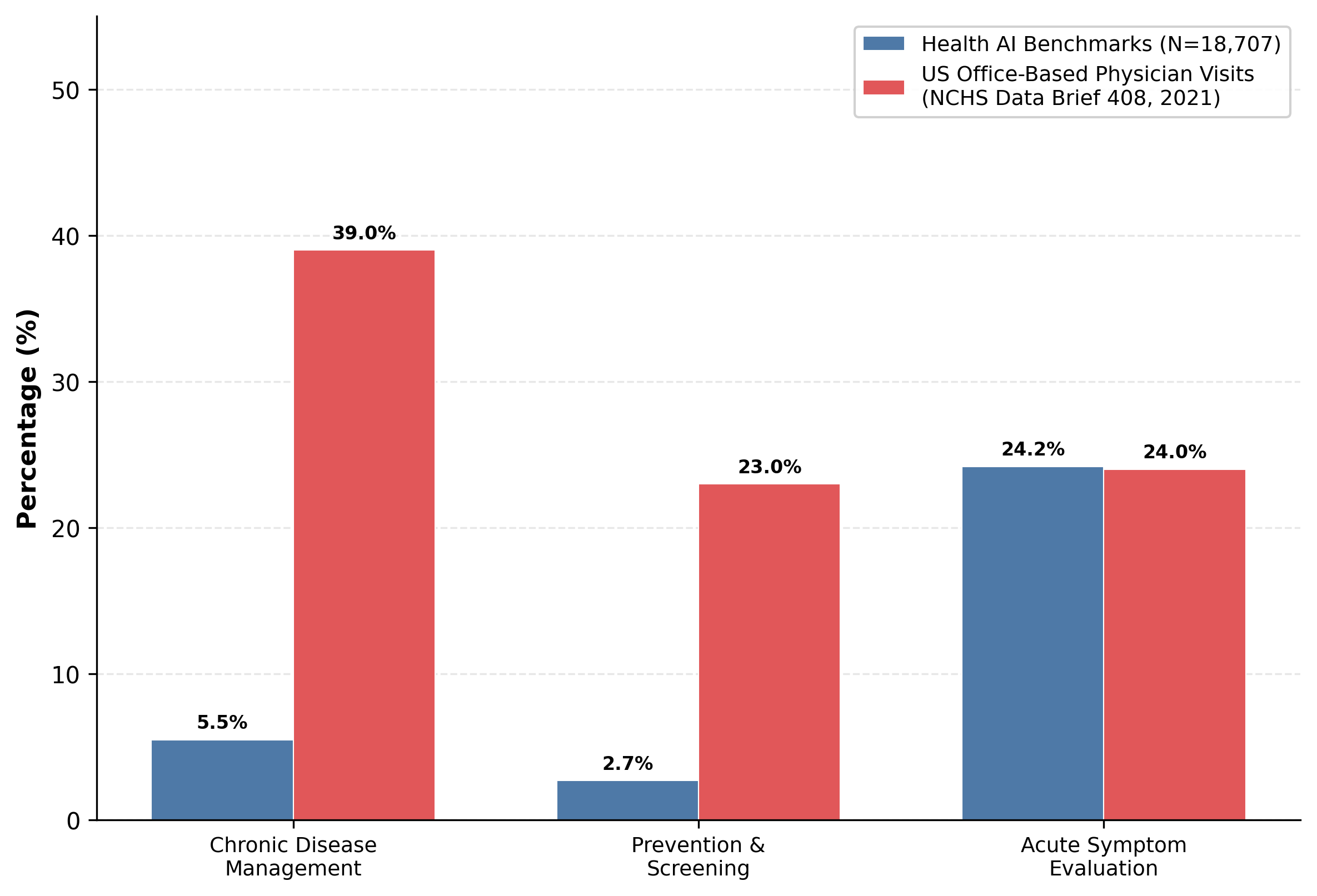}
\caption{Benchmark intent distribution compared to real-world office-based physician visit reasons (NCHS Data Brief No.\ 408, 2021). Chronic disease management accounts for 39\% of physician visits but only 5.5\% of benchmark queries---a 7-fold gap. Prevention and screening shows an even larger disparity (23\% vs.\ 2.7\%). Only acute symptom evaluation is proportionally represented in benchmarks (24.2\% vs.\ 24.0\%). Benchmark percentages do not sum to 100\% because additional intent categories (e.g., education, non-health) are not shown.}
\label{fig:chronic_care}
\end{figure}

This composition diverges sharply from real-world primary care (Figure~\ref{fig:chronic_care}). In a national survey of office-based physician visits, chronic conditions were the major reason for 39\% of all visits and preventive care accounted for 23\%~\cite{ashmanProductsDataBriefs2021}, proportions that have remained substantial even as overall primary care physician visits have declined~\cite{jettyEvaluationDeclinePrimary2025}. Acute symptom evaluation (new problems), by contrast, accounted for 24\% of visits. The benchmark distribution is effectively inverted: scenarios that dominate real-world clinical care---medication titration for hypertension, insulin management for diabetes, longitudinal monitoring of chronic conditions---are nearly absent from the datasets used to evaluate AI readiness for clinical deployment.

\subsection{Blind Spot 3: Clinical Document Interpretation}

As patients increasingly access their medical records through patient portals and bring clinical documents to AI systems for interpretation, the ability to parse and explain laboratory results, imaging reports, and other clinical artifacts becomes a core capability. Yet benchmarks provide minimal exposure to these scenarios. Explicit \textbf{laboratory results} appeared in only 5.2\% ($N=981$) of queries, \textbf{imaging reports} in 3.8\% ($N=702$), and raw \textbf{clinical artifacts}---text copied directly from electronic health records such as provider notes, discharge summaries, or pathology reports---in just 0.6\% ($N=110$).

While objective data of some form appeared in 42.3\% of the corpus, this figure was dominated by \textbf{wearable vitals} ($N=3,307$; 17.7\%), representing low-acuity signals such as step counts and sleep logs. The clinical documents that patients most commonly seek help interpreting---abnormal lab values, radiology findings, medication lists, specialist referral notes---were scarce. Moreover, queries containing multiple objective data types ($N=2,344$; 12.5\%), which most closely approximate the complexity of real clinical records, were concentrated almost entirely in Generation~2 (MedRedQA, 91.5\% of multi-type queries), where clinical context is pre-formulated in user narratives rather than presented as raw documents. This gap means that a model's ability to interpret actual clinical documents---the use case patients increasingly bring to AI chatbots---remains largely untested by public benchmarks.

\subsection{Blind Spot 4: Absence of Behavioral Health Crisis Scenarios}

\begin{figure}[H]
\centering
\includegraphics[width=1.0\textwidth]{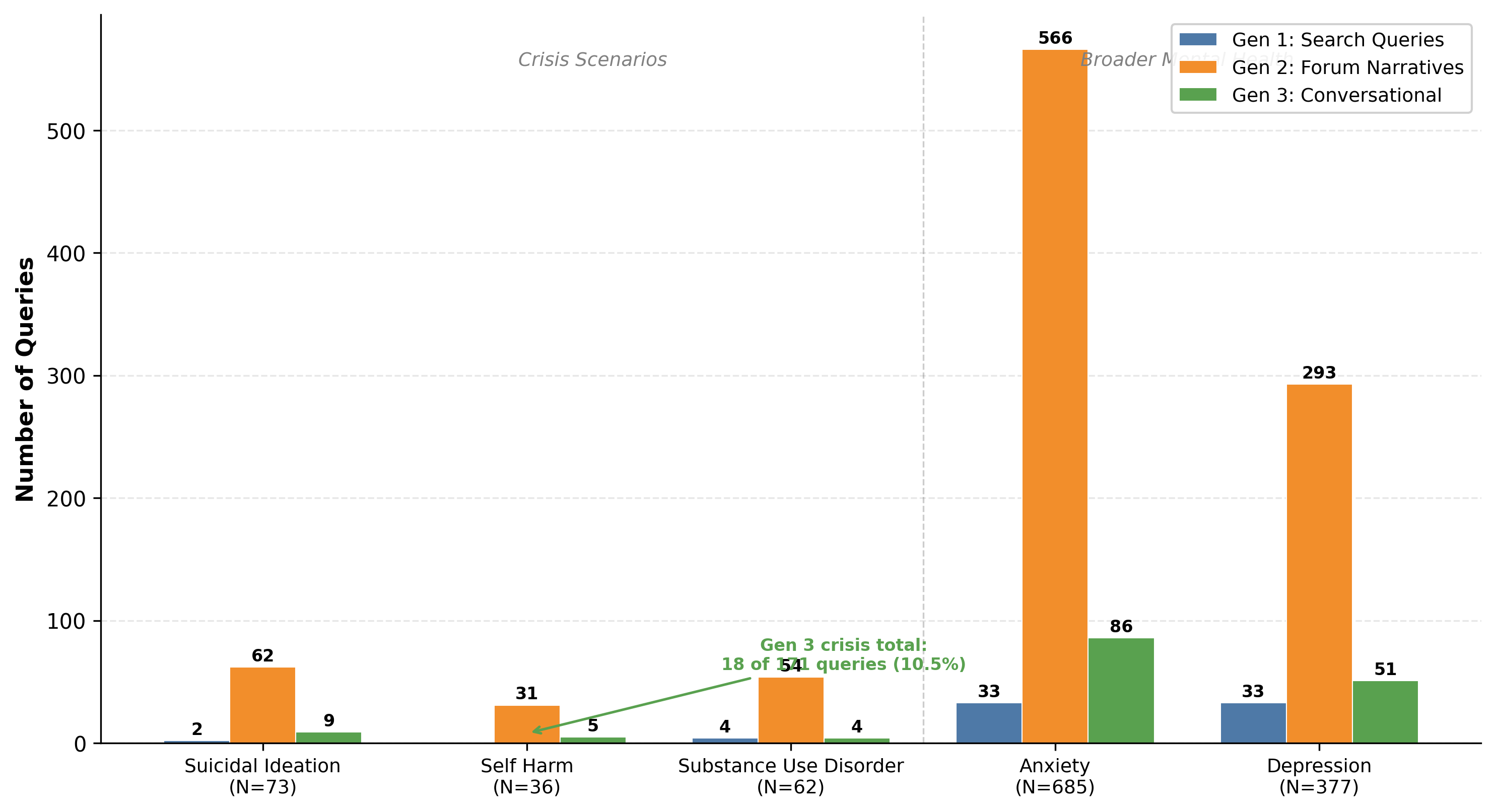}
\caption{Behavioral health crisis scenarios by benchmark generation. Crisis conditions (suicidal ideation, self-harm, substance use) were rare overall ($<$0.7\% of corpus) and concentrated almost entirely in Generation~2 forum narratives. Generation~3 conversational benchmarks---the modality where chatbots actually operate---contained fewer than 20 crisis queries combined, leaving crisis de-escalation in dialogue settings effectively untested. Broader mental health conditions (anxiety, depression) showed the same generational skew.}
\label{fig:behavioral_health}
\end{figure}

Despite the increasing deployment of AI chatbots for mental health support~\cite{metzAreAITherapy2025}, behavioral health crisis scenarios were effectively absent from the evaluation corpus. Queries involving \textbf{self-harm} ($N=36$; 0.19\%) and \textbf{suicidal ideation} ($N=73$; 0.39\%) combined represented less than 0.7\% of all queries (Figure~\ref{fig:behavioral_health}). These counts reflect explicit key condition mentions identified by the LLM tagger and differ from topic-level classifications in Table~S8.

Critically, the generation-level distribution reveals a deeper problem. Of the 73 suicidal ideation queries, 62 (84.9\%) appeared in Generation~2 forum narratives (MedRedQA), while only 9 (12.3\%) appeared in the conversational Generation~3 benchmarks where chatbots actually operate. The pattern was similar for self-harm: 31 of 36 queries (86.1\%) were in Generation~2, with only 5 (13.9\%) in Generation~3. Substance use disorder showed the same concentration: 54 of 62 queries (87.1\%) in Generation~2, just 4 (6.5\%) in Generation~3.

This distributional skew has a concrete clinical implication: the interactive modality in which chatbots actually encounter users in crisis---back-and-forth conversation---is the very modality left effectively untested for crisis de-escalation. Broader mental health conditions showed the same pattern: anxiety ($N=685$) and depression ($N=377$) were concentrated in Generation~2 (82.6\% and 77.7\%, respectively), with minimal representation in conversational benchmarks.

\subsection{Additional Compositional Findings}

Beyond the four primary blind spots, 72.7\% of queries were classified as low-risk, with high-risk scenarios concentrated almost entirely in Generation~2 (MedRedQA) and Generation~3 (HealthBench Main); excluding these datasets would eliminate nearly all exposure to clinically sensitive scenarios.

\section{Discussion}
% Discussion — standardized reporting proposals

Across 18,707 consumer health queries from six public benchmarks, we identified four systematic blind spots---in demographic representation, chronic disease management, clinical document interpretation, and behavioral health crisis scenarios---that collectively undermine the validity of current health AI evaluation. These are not marginal gaps. The conditions most likely to precipitate patient harm in AI-mediated care---inappropriate dosing for pediatric or geriatric patients, chronic disease management errors, misinterpretation of clinical documents, and suicidal crises in conversational settings---are precisely the scenarios that benchmarks fail to test. This compositional mismatch is analogous to validating a new therapeutic in a healthy population and extrapolating safety to patients with multimorbidity~\cite{rothwellExternalValidityRandomised2005}.

\paragraph{Clinical implications.} A natural question is whether models actually perform worse on underrepresented categories. Our claim is deliberately upstream of that question: current benchmarks provide no evaluation evidence for these categories at all. The problem is not demonstrated degradation but the absence of any basis to assess whether degradation exists. This echoes a long-standing concern that clinicians and patients who use machine-learning systems need to understand their limitations, including instances in which a model is not designed to generalize to a particular scenario~\cite{rajkomarMachineLearningMedicine2019}. Characterizing where evaluation evidence is absent is a necessary precondition for targeted performance analysis, and is itself the contribution. Each blind spot translates directly to deployment risk. The demographic skew toward a ``standard adult'' means that the populations with the greatest clinical complexity---children requiring weight-based dosing, older adults with polypharmacy and multimorbidity---are the populations for which evaluation evidence is weakest. Including fairness as a central consideration in how models are evaluated---not only how they are designed and deployed---is essential to ensuring that all patients benefit from this technology~\cite{rajkomarEnsuringFairnessMachine2018a}. The underrepresentation of chronic care---5.5\% in benchmarks versus 39\% of office-based physician visits nationally~\cite{ashmanProductsDataBriefs2021}---means that AI systems advising on medication management, disease monitoring, or preventive screening have minimal evaluation evidence for these core clinical functions. The scarcity of clinical document interpretation scenarios means that as patients increasingly use AI to understand their lab results, imaging reports, and discharge summaries, the accuracy of these interpretations has not been systematically evaluated. And the near-absence of behavioral health crises from conversational benchmarks means that chatbots marketed for mental health support~\cite{metzAreAITherapy2025} have not been evaluated for the highest-stakes interactions they will encounter.

When AI systems fail in these undertested scenarios, the cognitive burden of identifying and correcting errors falls back on physicians whose visit capacity is already declining~\cite{jettyEvaluationDeclinePrimary2025}. These gaps do not merely limit what we know about model performance; they create a false sense of security. A model achieving 90\% accuracy on a benchmark dominated by low-risk health education queries provides no assurance about its behavior when confronted with an elderly patient's complex medication list, a parent seeking to understand their child's lab results, or a suicidal adolescent in a chat interface.

\paragraph{Toward standardized query profiling.} To close this validity gap, we propose a reporting framework analogous to CONSORT guidelines for clinical trials~\cite{hopewellCONSORT2025Statement2025}. Critically, CONSORT does not require trials to demonstrate that excluded populations have worse outcomes---it requires trials to report who was enrolled so the field can reason about generalizability. We propose the same standard for benchmarks. A standardized ``Query Profile'' should accompany future health AI benchmarks, explicitly reporting five dimensions: (1)~\textbf{Clinical Topic Coverage} (balance of clinical domains); (2)~\textbf{Intent Distribution} (triage vs.\ education vs.\ chronic care); (3)~\textbf{Context Richness} (prevalence of clinical anchors); (4)~\textbf{Clinical Complexity} (acuity and risk levels); and (5)~\textbf{Data Integration} (presence of labs, vitals, or imaging). Such standardization would enable clinicians to determine whether a model has been evaluated on queries resembling their specific practice environment.

Beyond transparency, the community could benefit from a CONSORT-style extension process---convening informaticists, AI researchers, and journal editors to develop consensus standards for benchmark reporting. Over time, such standards could become embedded in publication and funding practices, much as CONSORT itself gradually shaped clinical trial reporting norms.

\paragraph{Open-source toolkit.} To lower the barrier for adoption, we have released an open-source Query Profiling toolkit. This automated pipeline takes any dataset in a standard format and generates the five-dimensional profile described above. The toolkit enables researchers to characterize new benchmarks without manual annotation, facilitating the systematic identification of compositional gaps.

\paragraph{Limitations.} Automated tagging using GPT-5.2 may miss nuances despite substantial cross-model agreement ($\kappa=0.77$) and high clinician-judged reasonableness (97\%). However, this trade-off is intentional: scalable, privacy-preserving profiling enables characterization of large corpora without manual annotation. Our taxonomy represents one pragmatic framework; alternative categorizations may reveal complementary insights. By design, this study characterizes the composition of evaluation evidence rather than model performance itself~\cite{tamFrameworkHumanEvaluation2024}; demonstrating where evidence is absent is logically prior to measuring performance within those gaps. External comparison data (NCHS Data Brief, WHO reports) provide approximate rather than exact benchmarks, as the clinical contexts are not directly equivalent. Finally, our taxonomy is cross-sectional; longitudinal benchmarks capturing evolving patient states will become essential as AI integrates deeper into care.

\subsection*{Conclusions}
Health AI benchmarks systematically omit the clinical scenarios most likely to cause patient harm: care for vulnerable populations, chronic disease management, clinical document interpretation, and behavioral health crises in conversational settings. Although individual benchmarks may be well-constructed and models perform well on many clinical tasks, aggregate performance scores on compositionally uncharacterized benchmarks do not reveal where evaluation evidence exists and where it is absent. We encourage the field to adopt routine query profiling---analogous to CONSORT reporting for clinical trials---as a standard practice to make the scope of evaluation transparent, enabling stakeholders to identify gaps and direct future benchmarking efforts where they are most needed.

\section*{Acknowledgments}
We thank Pooja Ramesh and Hywel Lo for technical assistance with the automated tagging system.

\subsection*{Conflicts of Interest}
All authors are employees of Apple. No external funding was received for this study.

% Bibliography from .bib file
\bibliographystyle{unsrtnat}
\bibliography{queryprofile}

% Tables and Figures
% Tables and Figures Section
\section*{Tables and Figures}
\label{sec:tables-figures}

\clearpage

%% ============================================================
%% SUPPLEMENTARY MATERIAL
%% ============================================================
\appendix
\section{Supplementary Material}
\label{app:supplement}

% Number tables and figures with S prefix for supplement
\setcounter{table}{0}
\renewcommand{\thetable}{S\arabic{table}}
\setcounter{figure}{0}
\renewcommand{\thefigure}{S\arabic{figure}}

\subsection{LLM Tagging Prompt}
\label{sec:llm-prompt}

The LLM tagging prompt (version 4.6) used with GPT-5.2 implements a deterministic classification framework with controlled vocabularies, priority ladders for intent disambiguation, and extensive validation rules. It classifies each query across 18 taxonomy dimensions including intent, topic, risk sensitivity, context depth, and objective data presence. The prompt includes few-shot examples and a validation checklist to ensure consistent output.

The full prompt is available in the project repository at \href{https://github.com/apple/ml-health-query-profiles}{github.com/apple/ml-health-query-profiles} (see \texttt{prompts/tagging\_prompt\_v4\_6.txt}).

\vspace{0.5em}
\noindent\textbf{Note on message format:} User messages were inserted in the following format:
\begin{itemize}
\item Single-turn queries: ``User: [message text]''
\item Multi-turn conversations: ``User: [message]$\langle$newline$\rangle$Assistant: [response]$\langle$newline$\rangle$User: [new message]...'' (where $\langle$newline$\rangle$ represents a line break in the actual prompt)
\end{itemize}

\clearpage
\subsection{Description of Rules-Based Tagging}
\label{sec:rules-based}

This appendix describes the rule-based approach used to annotate GoogleFitbit dataset queries with structured metadata tags. Unlike clinical benchmark datasets that used LLM-based tagging, the GoogleFitbit datasets employed deterministic programmatic classification due to their synthetic, highly-structured format.

\subsection{Rationale for Rule-Based Approach}

GoogleFitbit datasets consist of synthetic case studies with highly structured formats: demographic headers (age, gender), quantitative wearable data (sleep metrics, training load, heart rate variability), and health status indicators (BMI, conditions). This structural consistency enabled deterministic rule-based classification without LLMs.

Additionally, each case study was designed to be expanded into multiple individual prompts representing different analytical perspectives on the same case. This expansion step required programmatic logic to maintain consistency across related prompts derived from a single case.

\subsection{Datasets Tagged Programmatically}

Two GoogleFitbit datasets were processed using the programmatic approach:

\begin{itemize}
    \item \textbf{GoogleFitbit Fitness Cases}: 350 cases $\rightarrow$ 1,750 prompts (5$\times$ expansion)
    \item \textbf{GoogleFitbit Sleep Cases}: 507 cases $\rightarrow$ 1,521 prompts (3$\times$ expansion)
\end{itemize}

Processing occurred on October 3, 2025.

\subsection{Case-to-Prompt Expansion}

GoogleFitbit case studies were not analyzed as single queries but instead expanded into multiple analytical units to capture different facets of the wearable data:

\subsection{Fitness Cases (5 prompts per case)}

Each fitness case study was decomposed into five distinct analytical prompts as described in the original paper.

\begin{enumerate}
    \item \textbf{Demographics assessment}: Analysis of age, gender, and baseline characteristics
    \begin{itemize}
        \item Focus: Personal characteristics relevant to fitness coaching
    \end{itemize}

    \item \textbf{Training load analysis}: Exercise patterns and training volume
    \begin{itemize}
        \item Focus: Workout frequency, intensity, and progression
    \end{itemize}

    \item \textbf{Sleep metrics analysis}: Sleep quality and recovery indicators
    \begin{itemize}
        \item Focus: Sleep patterns and their impact on fitness performance
    \end{itemize}

    \item \textbf{Health metrics evaluation}: Physiological measurements
    \begin{itemize}
        \item Focus: Heart rate, heart rate variability, BMI, and other biomarkers
    \end{itemize}

    \item \textbf{Readiness assessment}: Training readiness and recovery status
    \begin{itemize}
        \item Focus: Muscle soreness, subjective readiness, and recovery recommendations
    \end{itemize}
\end{enumerate}

All fitness prompts were assigned the sub-intent classification \texttt{fitness\_exercise} under the top-level intent \texttt{general\_health\_advice}.

\subsection{Sleep Cases (3 prompts per case)}

Each sleep case study was decomposed into three analytical prompts:

\begin{enumerate}
    \item \textbf{Sleep insights}: Pattern identification from sleep logs
    \begin{itemize}
        \item Focus: Identifying trends and anomalies in sleep data
    \end{itemize}

    \item \textbf{Sleep etiology}: Root cause analysis of sleep issues
    \begin{itemize}
        \item Focus: Diagnosing potential factors affecting sleep quality
    \end{itemize}

    \item \textbf{Sleep recommendations}: Personalized intervention suggestions
    \begin{itemize}
        \item Focus: Actionable advice for improving sleep outcomes
    \end{itemize}
\end{enumerate}

All sleep prompts were assigned the sub-intent classification \texttt{sleep\_hygiene} under the top-level intent \texttt{general\_health\_advice}.

Each expanded prompt inherited core metadata from the parent case (demographics, detected conditions) while receiving the same intent classification reflecting the overall analytical focus on personalized wellness coaching.

\subsection{Rule-Based Classification Logic}

The programmatic tagger applied deterministic rules to extract features from structured case data and assign classification tags.

\subsection{Demographic Extraction}

Demographics were parsed from structured case headers using regex patterns:

\textbf{Age extraction and binning:}
\begin{itemize}
    \item Pattern matching: \texttt{[40-44]} (age range), \texttt{80+} (open-ended), \texttt{45 years old} (explicit)
    \item Range handling: Age ranges mapped to midpoint (e.g., [40-44] $\rightarrow$ 42)
    \item Binning rules:
    \begin{itemize}
        \item Age $<$ 18 $\rightarrow$ \texttt{pediatric}
        \item Age 18--64 $\rightarrow$ \texttt{adult}
        \item Age $\geq$ 65 $\rightarrow$ \texttt{adult\_65plus}
    \end{itemize}
\end{itemize}

\textbf{Gender extraction:}
\begin{itemize}
    \item Pattern matching from case headers: ``male,'' or ``gender: male''
    \item Extracted values: \texttt{male}, \texttt{female}
\end{itemize}

\subsection{Condition Detection}

The original rule-based tagging included keyword-based condition detection using patterns such as:

\textbf{Obesity detection (multi-method):}
\begin{itemize}
    \item Keyword match: ``obese'', ``obesity'', ``overweight''
    \item BMI threshold: BMI $\geq$ 30 extracted via regex (\texttt{bmi: 32.5})
    \item Either trigger results in \texttt{obesity} condition label
\end{itemize}

\textbf{Respiratory conditions:}
\begin{itemize}
    \item Keywords: ``asthma'', ``COPD'', ``breathing issues'', ``breathing problems''
    \item \textbf{Exclusion rule}: Mentions of ``Respiratory Rate'' (vital sign) do not trigger respiratory condition
\end{itemize}

\textbf{Other conditions (keyword-based):}
\begin{itemize}
    \item \textbf{Diabetes}: ``diabetes'', ``blood glucose'', ``insulin'', ``diabetic''
    \item \textbf{Hypertension}: ``hypertension'', ``high blood pressure''
    \item \textbf{Cardiovascular}: ``heart disease'', ``cardiovascular'', ``cardiac''
    \item \textbf{Joint issues}: ``arthritis'', ``joint pain'', ``knee pain'', ``back pain''
    \item \textbf{Metabolic syndrome}: ``metabolic syndrome'', ``cholesterol'', ``triglycerides''
\end{itemize}

\textbf{Note on data quality:} During validation, these regex-based condition detection rules were found to be unreliable, producing false positives (e.g., all 1,750 fitness cases were incorrectly tagged with ``obesity'' and ``respiratory'' conditions, and all 1,521 sleep cases with ``insomnia''). The \texttt{key\_conditions} field was therefore excluded from the final processed GoogleFitbit datasets to avoid contaminating downstream analyses with spurious condition labels.

\subsection{Fixed Tag Assignments}

All GoogleFitbit prompts received consistent tags reflecting their synthetic wearable data context. These assignments were hard-coded rather than inferred:

\begin{table}[h]
\centering
\small
\begin{tabular}{ll}
\toprule
\textbf{Tag Field} & \textbf{Assigned Value} \\
\midrule
\texttt{user\_type} & \texttt{consumer} \\
\texttt{conversation\_structure} & \texttt{single\_turn} \\
\texttt{intent\_top} & \texttt{general\_health\_advice} \\
\texttt{intent\_sub} & \texttt{fitness\_exercise} (fitness) or \texttt{sleep\_hygiene} (sleep) \\
\texttt{topic\_area\_path} & [\texttt{Holistic Health \& Wellness}, \texttt{Sleep \& Lifestyle}] \\
\texttt{specialty} & \texttt{sports\_medicine} (fitness) or \texttt{sleep\_medicine} (sleep) \\
\texttt{objective\_data} & \texttt{vitals\_wearable} \\
\texttt{context\_depth} & \texttt{high} \\
\texttt{setting} & \texttt{home} \\
\texttt{language\_complexity} & \texttt{lay} \\
\texttt{risk\_sensitivity} & \texttt{low} \\
\texttt{language} & \texttt{english} \\
\texttt{length\_detail} & \texttt{detailed} \\
\texttt{needs\_clarification} & \texttt{false} \\
\texttt{region} & \texttt{null} \\
\bottomrule
\end{tabular}
\caption{Fixed tag assignments for GoogleFitbit datasets reflecting wearable data context}
\label{tab:googlefitbit_fixed_tags}
\end{table}

\textbf{Rationale for fixed assignments:}
\begin{itemize}
    \item \texttt{consumer}: All case studies represent consumer health scenarios
    \item \texttt{single\_turn}: Synthetic cases have no prior conversation history
    \item \texttt{vitals\_wearable}: All cases include consumer device data (sleep tracking, heart rate, training metrics)
    \item \texttt{high} context depth: All cases provide detailed quantitative data including age, metrics over time, and health indicators
    \item \texttt{home}: Consumer wearable usage context
    \item \texttt{lay} complexity: Case descriptions written for general audiences
    \item \texttt{low} risk: Wellness and fitness optimization queries, not acute medical concerns
\end{itemize}

\clearpage
\subsection{Model Comparison Study}
\label{sec:model-comparison}

\subsection{Overview}
This section presents a systematic comparison of tagging outputs from two large language models---GPT-5.2 and Claude Opus~4.5---applied to the HealthBench Main dataset (N=4,967 matched records). The goal is to assess the reliability of LLM-derived tags by evaluating cross-model agreement across 21 structured dimensions.

Of the 5,000 queries in HealthBench Main, 33 (0.7\%) were not tagged by Opus~4.5 due to content filtering on queries involving sensitive clinical scenarios (e.g., pandemic-related questions, dangerous pathogens, terse symptom presentations). These queries were excluded from the comparison, yielding 4,967 matched pairs. This pattern is consistent with documented differences in content moderation policies between model providers and does not affect the validity of the comparison for the remaining queries.

\subsection{Methods}
Both models independently tagged the same queries using identical v4.6 tagging prompts with JSON output formatting. Agreement was assessed using:
\begin{itemize}
\item \textbf{Percent agreement}: Proportion of exact matches between models
\item \textbf{Cohen's $\kappa$}: Agreement corrected for chance, where $\kappa > 0.8$ indicates almost perfect agreement and $\kappa > 0.6$ indicates substantial agreement
\item \textbf{Chi-square tests}: Distribution equivalence with Cram\'er's V effect size
\end{itemize}

\subsection{Results}

\paragraph{Overall Agreement by Domain}
Across all 21 dimensions, models demonstrated strong agreement (Table~\ref{tab:S1-model-domain}). Context dimensions showed the highest concordance, followed by topic and intent dimensions.

\begin{table}[!ht]
\centering
\caption{Agreement by Query Profile Domain}
\label{tab:S1-model-domain}
\begin{tabular}{lccc}
\hline
\textbf{Domain} & \textbf{Dimensions} & \textbf{Avg Agreement} & \textbf{Avg $\kappa$} \\
\hline
Context & 16 & 92.6\% & 0.77 \\
Topic & 3 & 83.7\% & 0.79 \\
Intent & 2 & 78.5\% & 0.76 \\
\hline
\end{tabular}
\end{table}

\paragraph{Dimension-Level Agreement}
Table~\ref{tab:S2-model-dim} presents agreement metrics for each dimension.

\begin{table}[!ht]
\centering
\caption{Agreement Metrics by Dimension}
\label{tab:S2-model-dim}
\small
\begin{tabular}{llcc}
\hline
\textbf{Domain} & \textbf{Dimension} & \textbf{Agreement} & \textbf{$\kappa$} \\
\hline
\multicolumn{4}{l}{\textit{Context}} \\
& Conversation Structure & 100.0\% & 1.00 \\
& Language & 99.8\% & 0.99 \\
& Population & 98.6\% & 0.95 \\
& Raw Medical Text & 97.8\% & 0.71 \\
& Region & 96.9\% & 0.77 \\
& User Type & 96.9\% & 0.92 \\
& Personal Health Query & 95.3\% & 0.91 \\
& Language Complexity & 95.2\% & 0.88 \\
& Needs Clarification & 92.8\% & 0.17 \\
& Query Subject & 92.3\% & 0.89 \\
& Needs Personalization & 91.7\% & 0.77 \\
& Length/Detail & 90.3\% & 0.66 \\
& Context Depth & 85.9\% & 0.65 \\
& Risk Sensitivity & 85.9\% & 0.74 \\
& Objective Data & 83.3\% & 0.69 \\
& Setting & 79.1\% & 0.59 \\
\multicolumn{4}{l}{\textit{Topic}} \\
& Key Conditions & 91.7\% & 0.80 \\
& Specialty & 80.6\% & 0.79 \\
& Topic Area & 78.9\% & 0.79 \\
\multicolumn{4}{l}{\textit{Intent}} \\
& Top-Level Intent & 83.7\% & 0.81 \\
& Sub-Intent & 73.3\% & 0.71 \\
\hline
\end{tabular}
\end{table}

\paragraph{Distribution Comparisons}
Chi-square tests identified 5 dimensions with statistically significant distribution differences (p < 0.05 with Cram\'er\'s V > 0.1):
\begin{itemize}
\item Setting: $\chi^2$ = 403.6, V = 0.20
\item Needs Clarification: $\chi^2$ = 298.0, V = 0.17
\item Specialty: $\chi^2$ = 272.4, V = 0.17
\item Sub-Intent: $\chi^2$ = 204.8, V = 0.14
\item Context Depth: $\chi^2$ = 101.0, V = 0.10
\end{itemize}

\begin{figure}[!ht]
\centering
\includegraphics[width=\textwidth]{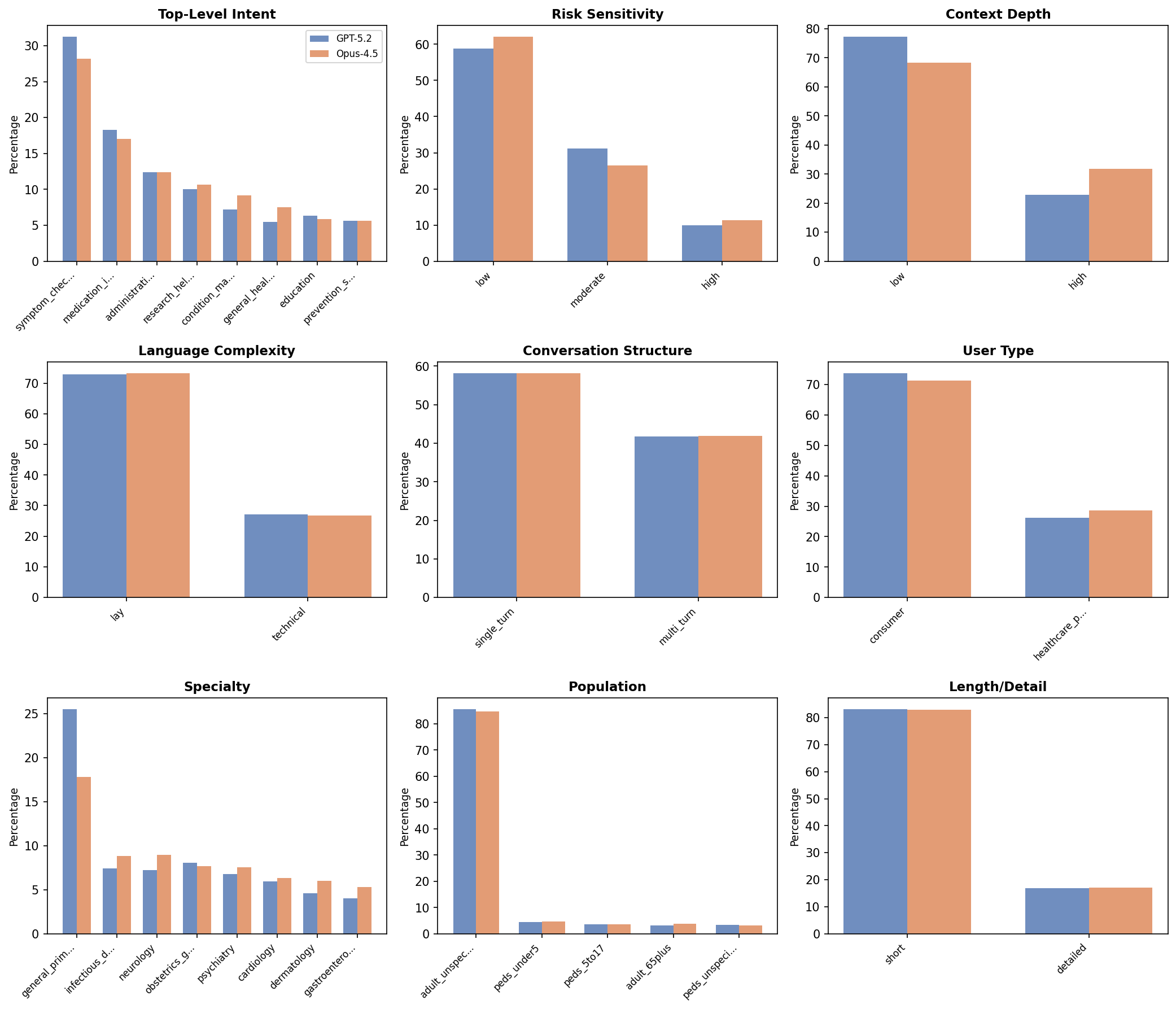}
\caption{Distribution comparison between GPT-5.2 and Opus-4.5 across key dimensions. Bars show percentage of queries assigned to each category.}
\label{fig:S1-model-distributions}
\end{figure}

\paragraph{Visual Comparison}
Figure~\ref{fig:S1-model-distributions} presents side-by-side distribution comparisons for key dimensions, demonstrating strong alignment between models across categorical values.

\begin{figure}[!ht]
\centering
\includegraphics[width=\textwidth]{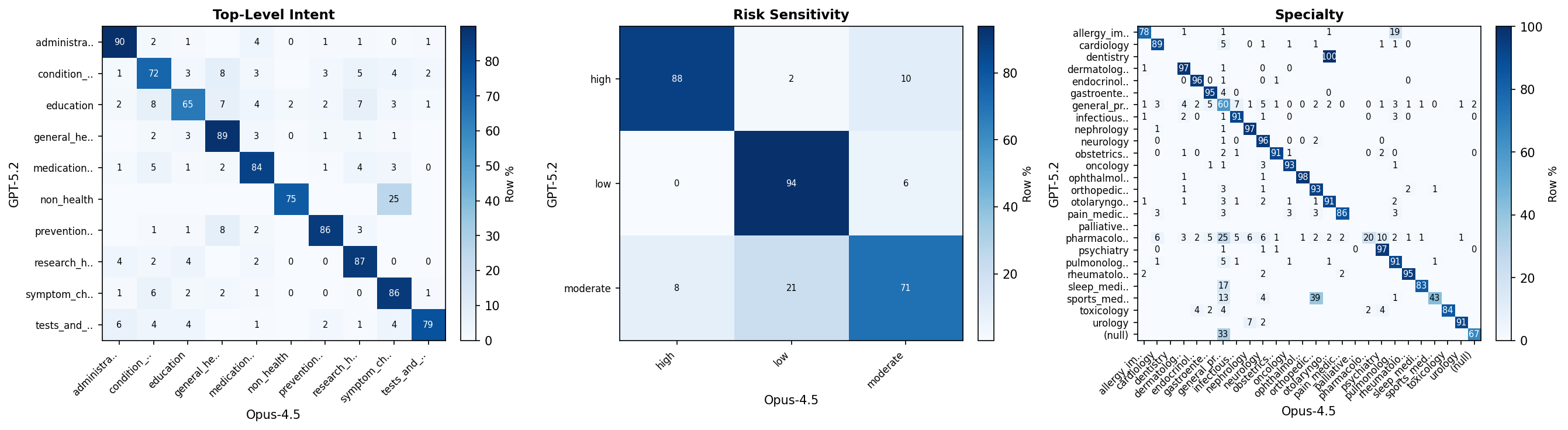}
\caption{Confusion matrices showing agreement patterns for intent, risk sensitivity, and specialty. Values show row-normalized percentages. Strong diagonal dominance indicates high agreement.}
\label{fig:S2-model-confusion}
\end{figure}

Figure~\ref{fig:S2-model-confusion} shows confusion matrices for dimensions with the most clinical relevance: intent classification, risk sensitivity, and specialty assignment.

\subsection{Discussion}

\paragraph{Implications for Tagging Validity}
The strong agreement between GPT-5.2 and Opus-4.5---two models with fundamentally different architectures and training approaches---provides compelling evidence for the validity of LLM-derived tags. Key findings include:
\begin{enumerate}
\item \textbf{Structural dimensions show near-perfect agreement}: Language, conversation structure, and user type achieved $>$95\% agreement, confirming these are objective, well-defined attributes.
\item \textbf{Aggregate distributions are statistically equivalent}: Chi-square tests revealed no meaningful distributional differences, indicating that population-level insights derived from either model would be consistent.
\item \textbf{Disagreements reflect genuine ambiguity}: Where models diverged (e.g., risk sensitivity boundaries), the disagreements occurred at category boundaries where human annotators would also exhibit variability.
\item \textbf{Cross-architecture validation}: Agreement between models trained by different organizations using different approaches provides stronger validity evidence than same-model reproducibility.
\end{enumerate}

\paragraph{Comparison to Human Inter-Rater Reliability}
The observed agreement levels ($\kappa$ = 0.6--0.9 across most dimensions) are comparable to or exceed typical human inter-rater reliability in medical annotation tasks, where $\kappa$ values of 0.4--0.7 are common for subjective clinical judgments. This suggests LLM tagging achieves human-level consistency while offering scalability advantages.

\paragraph{Limitations}
This comparison has limitations: (1) agreement was measured between models rather than against human gold standard; (2) both models used identical prompts, so shared prompt biases would not be detected; (3) analysis was limited to a single dataset. However, cross-model agreement provides meaningful validity evidence even without gold-standard comparison, as systematic biases are unlikely to be shared across independently developed models.

\clearpage
% Human Evaluation of LLM Tag Validity
% Two physician reviewers (A.R., L.H.P.) independently graded 80 stratified queries.

\subsection{Human Evaluation of Automated Tag Validity}
\label{app:human-eval}

To establish that the automated tagging produces clinically reasonable labels---not
merely that two LLMs agree with each other---we conducted a blinded human evaluation
study with two physician reviewers (a board-certified internist and an
ophthalmologist with health-AI expertise).

\subsubsection*{Rationale for Reasonableness as the Validation Criterion}

The cross-model agreement analysis (Section above) demonstrates that two
independently developed LLMs produce highly concordant tags
($\kappa = 0.77$), establishing \emph{reliability}.  However, reliability alone
does not guarantee \emph{accuracy}: both models could agree on the
wrong label.  A human evaluation is needed to close this gap.

We chose clinician-judged \emph{reasonableness} rather than agreement with a
human-generated gold standard for three reasons.  First, the Query Profile
taxonomy is a novel classification scheme with no pre-existing
human-annotated reference set.  Second, assigning queries to 16 structured
dimensions requires sustained attention to detailed definitions and
decision rules---a task better suited to automated systems that can apply
a multi-page prompt consistently across thousands of cases than to human
coders who would require extensive training and are susceptible to fatigue
and drift.  Constructing a human gold standard would therefore not
represent a higher-fidelity reference, but simply a different set of
annotator biases.  Third, the purpose of the tags is to provide reasonable
characterizations of health queries for downstream analysis; what matters
for this use case is not exact label match with any particular coder but
whether domain experts find the output clinically sensible.

Reasonableness validation---asking ``Would a clinician consider this
label appropriate?''---is therefore the most relevant and defensible
criterion.  It is analogous to validation approaches used in clinical
NLP, where system outputs are judged against clinician acceptability
rather than against a single reference annotation.

\subsubsection*{Sampling and Study Design}

From the 4{,}967 HealthBench Main records tagged by both GPT-5.2 and
Claude Opus~4.5, we restricted to English-language consumer queries
($n = 2{,}986$), then drew a stratified sample of 80 queries across the
three key taxonomic dimensions (risk sensitivity, top-level intent,
and Level-1 topic area).  The sample comprised two strata:

\begin{itemize}
  \item \textbf{Agreement stratum} ($n = 40$): queries where GPT-5.2 and
        Opus~4.5 assigned identical tags on all three dimensions.
        These were stratified proportionally by risk level
        (low 62\%, moderate 28\%, high 10\%) and then by intent category.
  \item \textbf{Disagreement stratum} ($n = 40$): queries where the models
        diverged on at least one dimension, stratified by which
        dimension(s) disagreed (12 risk-only, 14 intent-only,
        8 topic-only, 6 multi-dimension).
\end{itemize}

\subsubsection*{Annotation Procedure}

Each reviewer independently evaluated all 80 queries using a
self-contained browser-based tool.  Reviewers were blinded to model
identity throughout: agreement-stratum queries displayed a single tag
(the shared GPT-5.2 value), while disagreement-stratum queries
displayed two tags labeled only ``Model~A'' and ``Model~B,'' with
random assignment of GPT-5.2 versus Opus~4.5 to each label (fixed
seed per query).  An inline reference panel reproduced the exact
tag definitions from the LLM tagging prompt so that reviewers judged
tags against the same criteria used to generate them.

For each dimension on every query, reviewers provided:
\begin{enumerate}
  \item A 5-point Likert rating (1\,=\,strongly disagree to
        5\,=\,strongly agree) for the statement
        ``This tag is reasonable for this query.''
        On agreement-stratum queries, this applied to the consensus
        tag; on disagreement-stratum queries where models agreed on a
        given dimension, it applied to the shared value.
  \item For dimensions where models disagreed, a forced-choice
        preference (``Both reasonable, prefer A/B/no preference,''
        ``Only A reasonable,'' ``Only B reasonable,'' or ``Neither
        reasonable''), from which per-tag reasonableness was derived.
\end{enumerate}

\subsubsection*{Results}

\paragraph{Tag reasonableness.}
Across all 80 queries and both reviewers, the GPT-5.2 tag was judged
reasonable (Likert $\geq 4$) for 96.9\% of risk-sensitivity judgments,
96.8\% of intent judgments, and 97.1\% of topic-area judgments
(Table~\ref{tab:human-eval-accuracy}).  Mean Likert scores ranged from
4.84 to 4.90 out of~5.  Using a more lenient threshold ($\geq 3$),
reasonableness exceeded 97\% on all dimensions.  Per-reviewer accuracy
ranged from 93.8\% to 100\%, indicating consistent performance
across both clinicians.

\paragraph{Inter-rater agreement.}
Both reviewers rated the same tag as reasonable ($\geq 4$) on 92--95\%
of queries per dimension (Table~\ref{tab:human-eval-irr}).  Exact
Likert agreement (identical score) was 83--87\%, and agreement within
one point was 94--97\%.  Cohen's $\kappa$ was low (0.00--0.18) despite
high percentage agreement---a well-described artifact when prevalence
of agreement is very high and marginal distributions are skewed
(the ``kappa paradox''\cite{feinsteinHighAgreementLow1990}).
This pattern reflects genuine consensus, not measurement noise:
disagreements between reviewers were rare and always involved
borderline cases (e.g., Likert~3 versus~4), never diametrically
opposed judgments.

\paragraph{Model preference on disagreement cases.}
When the two LLMs produced different tags and reviewers chose between
blinded options, neither model was systematically preferred
(Table~\ref{tab:human-eval-preference}).  Across all three dimensions,
two-sided sign tests showed no significant preference for either GPT-5.2
or Opus~4.5 ($p = 0.36$--$0.83$).  Notably, zero query--dimension
pairs were rated ``Neither reasonable,'' meaning at least one model's
tag was always judged acceptable by both clinicians.

\paragraph{Estimated overall accuracy.}
Combining the agreement-stratum reasonableness rate with the
disagreement-stratum results (where at least one tag was always
acceptable), we estimate overall tagging accuracy of 89.6\%--91.7\%
across dimensions.  This lower bound is conservative: it counts
disagreement cases where one model was preferred as only 50\%
accurate, even though the preferred tag was judged reasonable.

% ── Tables ──

\begin{table}[!ht]
  \centering
  \caption{Clinician-judged reasonableness of GPT-5.2 tags (Likert $\geq 4$ = reasonable).
           Each cell pools both reviewers across 80 queries.}
  \label{tab:human-eval-accuracy}
  \small
  \begin{tabular}{lcccc}
    \toprule
    \textbf{Dimension} & $n$ & \textbf{Mean Likert} &
      \textbf{\% Reasonable} & \textbf{95\% CI} \\
    \midrule
    Risk Sensitivity     & 128 & 4.88 & 96.9\% & [92--99\%] \\
    Top-Level Intent     & 124 & 4.84 & 96.8\% & [92--99\%] \\
    Topic Area (Level 1) & 136 & 4.90 & 97.1\% & [93--99\%] \\
    \bottomrule
  \end{tabular}
\end{table}

\begin{table}[!ht]
  \centering
  \caption{Inter-rater agreement between two physician reviewers on the
           reasonable/not-reasonable binary ($\geq 4$ threshold).}
  \label{tab:human-eval-irr}
  \small
  \begin{tabular}{lcccc}
    \toprule
    \textbf{Dimension} & $n$ &
      \textbf{Both $\geq 4$} & \textbf{Exact Likert} &
      \textbf{Within 1 pt} \\
    \midrule
    Risk Sensitivity     & 64 & 93.8\% & 82.8\% & 93.8\% \\
    Top-Level Intent     & 62 & 95.2\% & 83.9\% & 96.8\% \\
    Topic Area (Level 1) & 68 & 94.1\% & 86.8\% & 94.1\% \\
    \bottomrule
  \end{tabular}
\end{table}

\begin{table}[!ht]
  \centering
  \caption{Model preference on disagreement cases (blinded).  ``GPT pref.'' and
           ``Opus pref.'' count queries where the respective model's tag was
           preferred; ``Both OK'' indicates both tags were acceptable with no
           preference.  Sign-test $p$-values test $H_0$: equal preference.}
  \label{tab:human-eval-preference}
  \small
  \begin{tabular}{lccccc}
    \toprule
    \textbf{Dimension} & \textbf{GPT pref.} & \textbf{Opus pref.} &
      \textbf{Both OK} & \textbf{Neither} & \textbf{Sign $p$} \\
    \midrule
    Risk Sensitivity     & 18 & 12 & 2 & 0 & 0.362 \\
    Top-Level Intent     & 17 & 14 & 5 & 0 & 0.720 \\
    Topic Area (Level 1) & 10 & 12 & 2 & 0 & 0.832 \\
    \bottomrule
  \end{tabular}
\end{table}

\subsubsection*{Interpretation}

These results support three conclusions relevant to the validity of the
automated tagging framework:

\begin{enumerate}
  \item \textbf{High accuracy:} Physician reviewers judged the GPT-5.2
        tag as reasonable in $\geq$\,96\% of cases across all three
        dimensions, with mean Likert scores near the ceiling (4.84--4.90/5).
        This establishes that the automated tags are not merely
        internally consistent (as shown by the 90\% cross-model
        agreement, $\kappa = 0.77$) but also \emph{externally valid}
        against independent clinical judgment.

  \item \textbf{No model dominance:} Neither GPT-5.2 nor Opus~4.5
        was systematically preferred by clinicians on disagreement
        cases ($p > 0.35$ on all dimensions), and in every case at
        least one model produced an acceptable tag.  This suggests
        that model disagreements typically reflect legitimate
        ambiguity in the tagging task rather than systematic error by
        either model.

  \item \textbf{Consistent across reviewers:} Both physicians
        independently arrived at high reasonableness ratings (93.8--100\%
        per reviewer per dimension) with 94--97\% agreement within one
        Likert point, confirming that the evaluation reflects genuine
        tag quality rather than idiosyncratic judgment.
\end{enumerate}

\subsubsection*{Potential Limitations and Mitigations}

The high Likert scores (mean 4.84--4.90/5) could be interpreted as a
ceiling effect suggesting that the evaluation was too easy.  Several
design features mitigate this concern.  First, the sample was
deliberately enriched with difficult cases: half of all queries came from
the disagreement stratum, where at least one LLM produced a tag that
differed from the other.  Despite this enrichment, clinicians still
judged the tags as reasonable in $>$96\% of cases, indicating that the
high scores reflect genuine tag quality, not task ease.  Second, the
5-point Likert scale provided sufficient resolution to detect
dissatisfaction---scores of 1--3 were available and were used (3--7\% of
judgments per dimension)---so the instrument was not insensitive to poor
tags.  Third, reviewers were not asked to generate their own labels
(which could introduce anchoring bias toward agreement); they evaluated
blinded labels against explicit definitions, preserving independence.

With two reviewers, the study was not powered to detect subtle differences
in inter-rater reliability across dimensions.  However, the primary claim
does not rest on inter-rater statistics: it rests on the consistently
high absolute reasonableness rate observed independently by both
physicians, which is robust to sample size.

Finally, we note that the two reviewers bring complementary
clinical perspectives (internal medicine and ophthalmology/health-AI),
increasing confidence that the tags are reasonable across clinical
viewpoints rather than reflecting the expertise of a single specialty.

\subsubsection*{Summary}

Taken together, the cross-model agreement analysis and this human
evaluation provide complementary evidence for tagging validity:
the former demonstrates \emph{reliability} (reproducibility across
models), while the latter demonstrates \emph{accuracy} (alignment with
clinical judgment).  The validation strategy is appropriate for a novel
taxonomy where no pre-existing gold standard exists: rather than
comparing against an imperfect human reference, we directly assessed
whether the automated output meets the clinically meaningful bar of
reasonableness.  The combination supports the use of LLM-based
tagging as a scalable and valid mechanism for characterizing
health-related queries.

\clearpage
\subsection{Supplementary Figures}
\label{sec:supplementary-figures}

\begin{figure}[ht]
\centering
\includegraphics[width=1.0\textwidth]{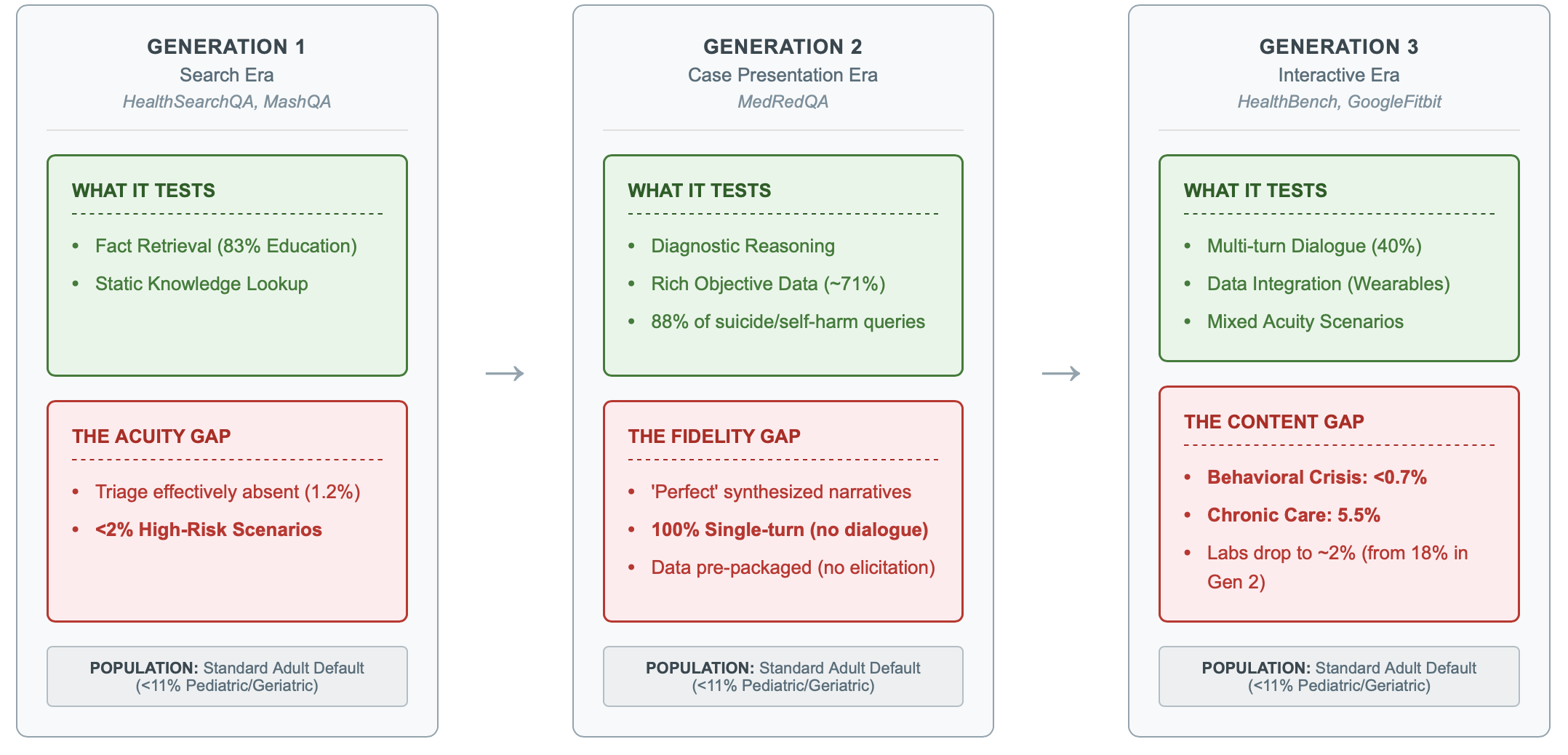}
\caption{\textbf{The Evolution of the Validity Gap.} Each generation of benchmarks advances structural capabilities but retains specific validity gaps. \textbf{Generation 1} evaluates fact retrieval but lacks acuity for triage (1.2\% triage intent, $<$2\% high-risk). \textbf{Generation 2} introduces clinical reasoning with rich objective data but relies on synthesized, static narratives (100\% single-turn) where data is pre-packaged rather than elicited. \textbf{Generation 3} achieves interactivity but suffers from sparse clinical content: $<$0.7\% behavioral crisis, 5.5\% chronic care, and clinical data density drops compared to Generation 2. Critically, demographic bias persists across all generations: pediatric and geriatric populations represent $<$11\% of queries, validating models on a ``standard adult default.''}
\label{fig:generations}
\end{figure}

\clearpage

\begin{figure}[h]
\centering
\includegraphics[width=1.0\textwidth]{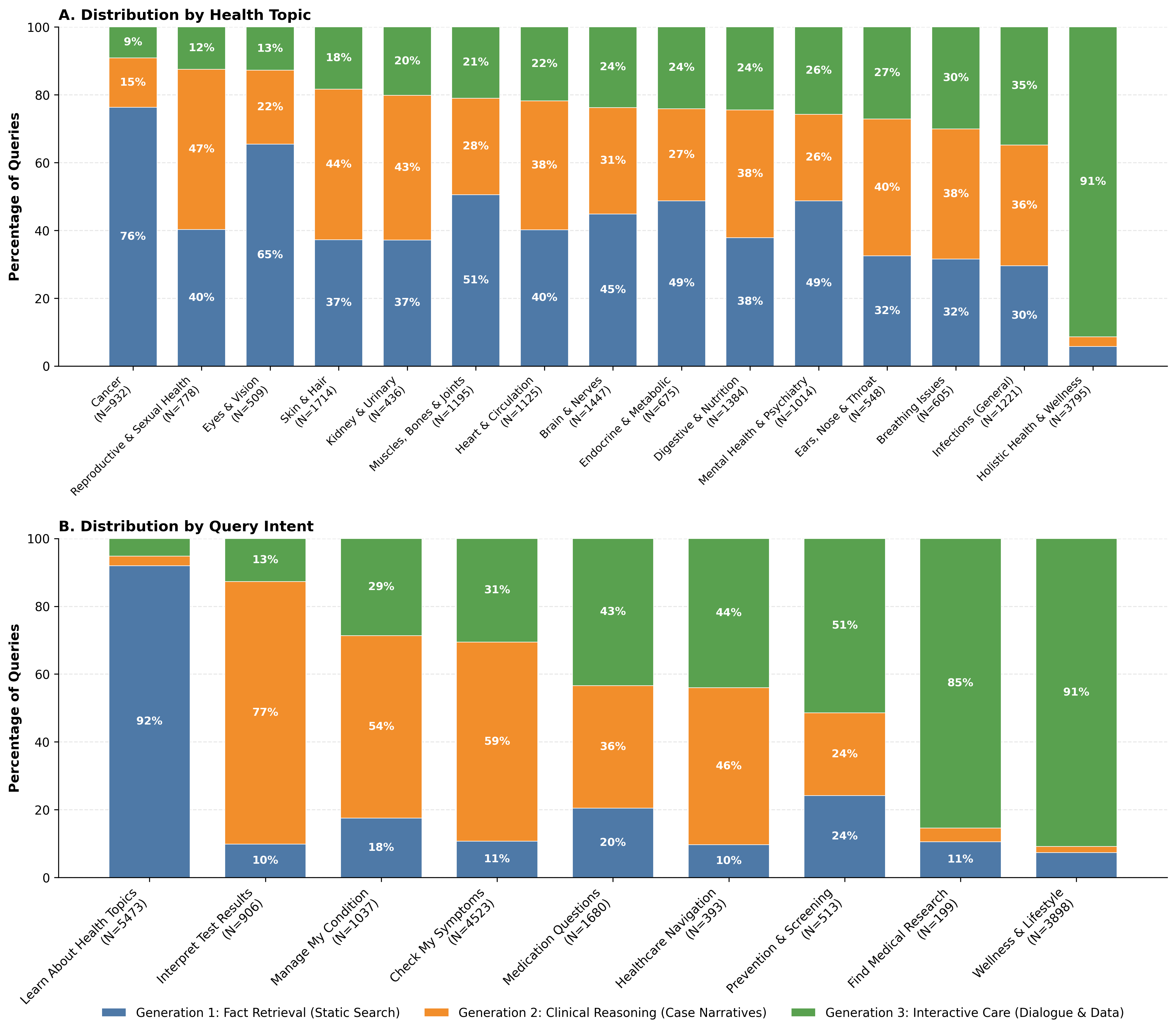}
\caption{Distribution of health topics and query intents across three generations of benchmarks. \textbf{(A)} Health topics: Generation 1 (Search Era) queries span common symptoms and general health concerns. Generation 2 (Case Presentation Era) concentrates in internal medicine subspecialties with higher clinical complexity. Generation 3 (Interactive \& Data Era) shows bimodal distribution: HealthBench Main covers diverse acute and chronic conditions, while GoogleFitbit datasets focus predominantly on wellness, sleep, and fitness tracking. \textbf{(B)} Query intents: Generation 1 benchmarks are dominated by queries to learn about health topics (92\%), while Generation 3 benchmarks capture broader intent diversity including wellness and lifestyle guidance, medical research queries, and prevention and screening.}
\label{fig:generation_distribution}
\end{figure}

\clearpage
\subsection{Supplementary Tables}
\label{sec:supplementary-tables}

\afterpage{%
\clearpage
\begin{landscape}
\small
\begin{longtable}{p{4cm}p{2.5cm}p{2.5cm}p{2.5cm}p{2.5cm}p{2.5cm}p{2.5cm}p{2cm}}
\caption{Baseline Structural and Clinical Characteristics of the Analytic Cohort (N=18,707 Queries)} \label{tab:baseline_characteristics} \\
\toprule
\textbf{Characteristic} & \textbf{HealthSearchQA} & \textbf{MashQA Test} & \textbf{MedRedQA} & \textbf{HealthBench} & \textbf{GoogleFitbit} & \textbf{GoogleFitbit} & \textbf{Total} \\
 &  &  & \textbf{Test} & \textbf{Main} & \textbf{Sleep} & \textbf{Fitness} & \\
\midrule
\endfirsthead
\multicolumn{8}{c}{\tablename\ \thetable\ -- \emph{Continued from previous page}} \\
\toprule
\textbf{Characteristic} & \textbf{HealthSearchQA} & \textbf{MashQA Test} & \textbf{MedRedQA} & \textbf{HealthBench} & \textbf{GoogleFitbit} & \textbf{GoogleFitbit} & \textbf{Total} \\
 &  &  & \textbf{Test} & \textbf{Main} & \textbf{Sleep} & \textbf{Fitness} & \\
\midrule
\endhead
\midrule
\multicolumn{8}{r}{\emph{Continued on next page}} \\
\endfoot
\bottomrule
\endlastfoot

\textbf{Total Records} & 3173 & 3490 & 5081 & 3692 & 1521 & 1750 & \textbf{18707} \\
\multicolumn{8}{l}{\textbf{Population}} \\
Adult (age unspecified), n (\%) & 3104 (97.8\%) & 3297 (94.5\%) & 4394 (86.5\%) & 3249 (88.0\%) & 1092 (71.8\%) & 1520 (86.9\%) & \textbf{16656 (89.0\%)} \\
Adult (65+ years), n (\%) & 2 (0.1\%) & 13 (0.4\%) & 117 (2.3\%) & 70 (1.9\%) & 429 (28.2\%) & 230 (13.1\%) & \textbf{861 (4.6\%)} \\
Pediatric (age unspecified), n (\%) & 40 (1.3\%) & 115 (3.3\%) & 5 (0.1\%) & 121 (3.3\%) & 0 (0.0\%) & 0 (0.0\%) & \textbf{281 (1.5\%)} \\
Pediatric (under 5 years), n (\%) & 26 (0.8\%) & 57 (1.6\%) & 97 (1.9\%) & 159 (4.3\%) & 0 (0.0\%) & 0 (0.0\%) & \textbf{339 (1.8\%)} \\
Pediatric (5-17 years), n (\%) & 1 (0.0\%) & 8 (0.2\%) & 468 (9.2\%) & 93 (2.5\%) & 0 (0.0\%) & 0 (0.0\%) & \textbf{570 (3.0\%)} \\
\addlinespace
\multicolumn{8}{l}{\textbf{Prompt Length (words)}} \\
Median (IQR) & 7 (5-8) & 8 (6-10) & 150 (98-241) & 16 (9-30) & 659 (514-2326) & 414 (44-434) & \textbf{209 (5-232)} \\
\addlinespace
\multicolumn{8}{l}{\textbf{Language}} \\
English, n (\%) & 3173 (100.0\%) & 3490 (100.0\%) & 5081 (100.0\%) & 3010 (81.5\%) & 1521 (100.0\%) & 1750 (100.0\%) & \textbf{18025 (96.4\%)} \\
\addlinespace
\multicolumn{8}{l}{\textbf{Conversation Structure}} \\
Single-turn, n (\%) & 3173 (100.0\%) & 3490 (100.0\%) & 5072 (99.8\%) & 2193 (59.4\%) & 1521 (100.0\%) & 1750 (100.0\%) & \textbf{17199 (91.9\%)} \\
Multi-turn, n (\%) & 0 (0.0\%) & 0 (0.0\%) & 9 (0.2\%) & 1499 (40.6\%) & 0 (0.0\%) & 0 (0.0\%) & \textbf{1508 (8.1\%)} \\
\addlinespace
\multicolumn{8}{l}{\textbf{Risk Sensitivity}} \\
Low, n (\%) & 3093 (97.5\%) & 3441 (98.6\%) & 1639 (32.3\%) & 2161 (58.5\%) & 1521 (100.0\%) & 1750 (100.0\%) & \textbf{13605 (72.7\%)} \\
Moderate, n (\%) & 70 (2.2\%) & 44 (1.3\%) & 2942 (57.9\%) & 1189 (32.2\%) & 0 (0.0\%) & 0 (0.0\%) & \textbf{4245 (22.7\%)} \\
High, n (\%) & 10 (0.3\%) & 5 (0.1\%) & 500 (9.8\%) & 342 (9.3\%) & 0 (0.0\%) & 0 (0.0\%) & \textbf{857 (4.6\%)} \\
\addlinespace
\multicolumn{8}{l}{\textbf{Language Complexity}} \\
Lay language, n (\%) & 3069 (96.7\%) & 3157 (90.5\%) & 4461 (87.8\%) & 3470 (94.0\%) & 1521 (100.0\%) & 1750 (100.0\%) & \textbf{17428 (93.2\%)} \\
Technical language, n (\%) & 104 (3.3\%) & 333 (9.5\%) & 620 (12.2\%) & 222 (6.0\%) & 0 (0.0\%) & 0 (0.0\%) & \textbf{1279 (6.8\%)} \\
\addlinespace
\multicolumn{8}{l}{\textbf{Objective Data Present}} \\
None, n (\%) & 3165 (99.7\%) & 3357 (96.2\%) & 1435 (28.2\%) & 2829 (76.6\%) & 0 (0.0\%) & 0 (0.0\%) & \textbf{10786 (57.7\%)} \\
Wearable vitals, n (\%) & 0 (0.0\%) & 0 (0.0\%) & 35 (0.7\%) & 1 (0.0\%) & 1521 (100.0\%) & 1750 (100.0\%) & \textbf{3307 (17.7\%)} \\
Diagnoses, n (\%) & 5 (0.2\%) & 110 (3.2\%) & 2230 (43.9\%) & 546 (14.8\%) & 0 (0.0\%) & 0 (0.0\%) & \textbf{2891 (15.5\%)} \\
Medications, n (\%) & 0 (0.0\%) & 21 (0.6\%) & 2044 (40.2\%) & 257 (7.0\%) & 0 (0.0\%) & 0 (0.0\%) & \textbf{2322 (12.4\%)} \\
Procedures, n (\%) & 0 (0.0\%) & 15 (0.4\%) & 879 (17.3\%) & 99 (2.7\%) & 0 (0.0\%) & 0 (0.0\%) & \textbf{993 (5.3\%)} \\
Laboratory results, n (\%) & 0 (0.0\%) & 1 (0.0\%) & 898 (17.7\%) & 82 (2.2\%) & 0 (0.0\%) & 0 (0.0\%) & \textbf{981 (5.2\%)} \\
Imaging, n (\%) & 0 (0.0\%) & 0 (0.0\%) & 659 (13.0\%) & 43 (1.2\%) & 0 (0.0\%) & 0 (0.0\%) & \textbf{702 (3.8\%)} \\
Basic vitals, n (\%) & 3 (0.1\%) & 0 (0.0\%) & 383 (7.5\%) & 66 (1.8\%) & 0 (0.0\%) & 0 (0.0\%) & \textbf{452 (2.4\%)} \\
Clinical artifacts, n (\%) & 0 (0.0\%) & 0 (0.0\%) & 93 (1.8\%) & 17 (0.5\%) & 0 (0.0\%) & 0 (0.0\%) & \textbf{110 (0.6\%)} \\
\addlinespace
\multicolumn{8}{l}{\textbf{Query Subject}} \\
Self, n (\%) & 218 (6.9\%) & 223 (6.4\%) & 4327 (85.2\%) & 2247 (60.9\%) & 1521 (100.0\%) & 1750 (100.0\%) & \textbf{10286 (55.0\%)} \\
General, n (\%) & 2944 (92.8\%) & 3215 (92.1\%) & 85 (1.7\%) & 870 (23.6\%) & 0 (0.0\%) & 0 (0.0\%) & \textbf{7114 (38.0\%)} \\
Child, n (\%) & 11 (0.3\%) & 52 (1.5\%) & 160 (3.1\%) & 309 (8.4\%) & 0 (0.0\%) & 0 (0.0\%) & \textbf{532 (2.8\%)} \\
Parent, n (\%) & 0 (0.0\%) & 0 (0.0\%) & 186 (3.7\%) & 75 (2.0\%) & 0 (0.0\%) & 0 (0.0\%) & \textbf{261 (1.4\%)} \\
Partner, n (\%) & 0 (0.0\%) & 0 (0.0\%) & 158 (3.1\%) & 23 (0.6\%) & 0 (0.0\%) & 0 (0.0\%) & \textbf{181 (1.0\%)} \\
Other, n (\%) & 0 (0.0\%) & 0 (0.0\%) & 165 (3.2\%) & 168 (4.6\%) & 0 (0.0\%) & 0 (0.0\%) & \textbf{333 (1.8\%)} \\

\end{longtable}
\end{landscape}
}

\vspace{0.5cm}
{\small \emph{Values are presented as n (\%) for categorical variables and median (IQR) for continuous variables.}}

\clearpage

\afterpage{%
\clearpage
\begin{landscape}
\small
\begin{longtable}{p{3.5cm}p{2.8cm}p{2.8cm}p{2.8cm}p{2.8cm}p{2.8cm}p{2.8cm}p{2.2cm}}
\caption{Distribution of Clinical Intents Across 'Generations' of Health AI Benchmarks} \label{tab:intents} \\
\toprule
\textbf{Intent/Sub-Intent} & \textbf{HealthSearchQA} & \textbf{MashQA Test} & \textbf{MedRedQA} & \textbf{HealthBench} & \textbf{GoogleFitbit} & \textbf{GoogleFitbit} & \textbf{Total} \\
 &  &  & \textbf{Test} & \textbf{Main} & \textbf{Sleep} & \textbf{Fitness} & \\
\addlinespace[0.5ex]
\midrule
\endfirsthead
\multicolumn{8}{c}{\tablename\ \thetable\ -- \emph{Continued from previous page}} \\
\toprule
\textbf{Intent/Sub-Intent} & \textbf{HealthSearchQA} & \textbf{MashQA Test} & \textbf{MedRedQA} & \textbf{HealthBench} & \textbf{GoogleFitbit} & \textbf{GoogleFitbit} & \textbf{Total} \\
 &  &  & \textbf{Test} & \textbf{Main} & \textbf{Sleep} & \textbf{Fitness} & \\
\addlinespace[0.5ex]
\midrule
\endhead
\midrule
\multicolumn{8}{r}{\emph{Continued on next page}} \\
\endfoot
\bottomrule
\endlastfoot

\textbf{Learn About Health Topics} & 2645 (83.4\%) & 2389 (68.5\%) & 154 (3.0\%) & 285 (7.7\%) & 0 (0.0\%) & 0 (0.0\%) & \textbf{5473 (29.3\%)} \\
\quad Education Explainer & 2635 (83.0\%) & 2250 (64.5\%) & 149 (2.9\%) & 280 (7.6\%) & 0 (0.0\%) & 0 (0.0\%) & 5314 (28.4\%) \\
\quad Basic Science & 10 (0.3\%) & 139 (4.0\%) & 5 (0.1\%) & 5 (0.1\%) & 0 (0.0\%) & 0 (0.0\%) & 159 (0.8\%) \\
\addlinespace[0.5ex]
\textbf{Check My Symptoms} & 380 (12.0\%) & 103 (3.0\%) & 2656 (52.3\%) & 1384 (37.5\%) & 0 (0.0\%) & 0 (0.0\%) & \textbf{4523 (24.2\%)} \\
\quad Management Plan & 235 (7.4\%) & 31 (0.9\%) & 774 (15.2\%) & 681 (18.4\%) & 0 (0.0\%) & 0 (0.0\%) & 1721 (9.2\%) \\
\quad Differential Diagnosis & 107 (3.4\%) & 20 (0.6\%) & 1235 (24.3\%) & 311 (8.4\%) & 0 (0.0\%) & 0 (0.0\%) & 1673 (8.9\%) \\
\quad Triage Disposition & 38 (1.2\%) & 52 (1.5\%) & 647 (12.7\%) & 392 (10.6\%) & 0 (0.0\%) & 0 (0.0\%) & 1129 (6.0\%) \\
\addlinespace[0.5ex]
\textbf{Wellness \& Lifestyle} & 47 (1.5\%) & 240 (6.9\%) & 70 (1.4\%) & 270 (7.3\%) & 1521 (100.0\%) & 1750 (100.0\%) & \textbf{3898 (20.8\%)} \\
\quad Fitness Exercise & 1 (0.0\%) & 40 (1.1\%) & 6 (0.1\%) & 40 (1.1\%) & 0 (0.0\%) & 1750 (100.0\%) & 1837 (9.8\%) \\
\quad Sleep Hygiene & 8 (0.3\%) & 6 (0.2\%) & 4 (0.1\%) & 20 (0.5\%) & 1521 (100.0\%) & 0 (0.0\%) & 1559 (8.3\%) \\
\quad Nutrition Diet & 19 (0.6\%) & 72 (2.1\%) & 25 (0.5\%) & 115 (3.1\%) & 0 (0.0\%) & 0 (0.0\%) & 231 (1.2\%) \\
\quad Supplements Nutraceuticals & 2 (0.1\%) & 64 (1.8\%) & 15 (0.3\%) & 60 (1.6\%) & 0 (0.0\%) & 0 (0.0\%) & 141 (0.8\%) \\
\quad Stress Selfcare & 7 (0.2\%) & 40 (1.1\%) & 8 (0.2\%) & 6 (0.2\%) & 0 (0.0\%) & 0 (0.0\%) & 61 (0.3\%) \\
\quad Cosmeceuticals Topicals & 10 (0.3\%) & 10 (0.3\%) & 9 (0.2\%) & 24 (0.7\%) & 0 (0.0\%) & 0 (0.0\%) & 53 (0.3\%) \\
\quad Nutrition Facts Lookup & 0 (0.0\%) & 8 (0.2\%) & 2 (0.0\%) & 5 (0.1\%) & 0 (0.0\%) & 0 (0.0\%) & 15 (0.1\%) \\
\quad Lifestyle Prevention & 0 (0.0\%) & 0 (0.0\%) & 1 (0.0\%) & 0 (0.0\%) & 0 (0.0\%) & 0 (0.0\%) & 1 (0.0\%) \\
\addlinespace[0.5ex]
\textbf{Medication Questions} & 11 (0.3\%) & 332 (9.5\%) & 608 (12.0\%) & 729 (19.7\%) & 0 (0.0\%) & 0 (0.0\%) & \textbf{1680 (9.0\%)} \\
\quad Selection & 8 (0.3\%) & 112 (3.2\%) & 134 (2.6\%) & 443 (12.0\%) & 0 (0.0\%) & 0 (0.0\%) & 697 (3.7\%) \\
\quad Side Effects & 2 (0.1\%) & 157 (4.5\%) & 240 (4.7\%) & 72 (2.0\%) & 0 (0.0\%) & 0 (0.0\%) & 471 (2.5\%) \\
\quad Dosing & 0 (0.0\%) & 45 (1.3\%) & 99 (1.9\%) & 170 (4.6\%) & 0 (0.0\%) & 0 (0.0\%) & 314 (1.7\%) \\
\quad Interactions & 1 (0.0\%) & 8 (0.2\%) & 119 (2.3\%) & 26 (0.7\%) & 0 (0.0\%) & 0 (0.0\%) & 154 (0.8\%) \\
\quad Safety Preg Lact & 0 (0.0\%) & 10 (0.3\%) & 16 (0.3\%) & 18 (0.5\%) & 0 (0.0\%) & 0 (0.0\%) & 44 (0.2\%) \\
\addlinespace[0.5ex]
\textbf{Manage My Condition} & 25 (0.8\%) & 157 (4.5\%) & 558 (11.0\%) & 297 (8.0\%) & 0 (0.0\%) & 0 (0.0\%) & \textbf{1037 (5.5\%)} \\
\quad Chronic Care Support & 21 (0.7\%) & 141 (4.0\%) & 209 (4.1\%) & 227 (6.1\%) & 0 (0.0\%) & 0 (0.0\%) & 598 (3.2\%) \\
\quad Risk Prognosis & 0 (0.0\%) & 10 (0.3\%) & 212 (4.2\%) & 44 (1.2\%) & 0 (0.0\%) & 0 (0.0\%) & 266 (1.4\%) \\
\quad Acute Flare Management & 4 (0.1\%) & 6 (0.2\%) & 137 (2.7\%) & 26 (0.7\%) & 0 (0.0\%) & 0 (0.0\%) & 173 (0.9\%) \\
\addlinespace[0.5ex]
\textbf{Interpret Test Results} & 7 (0.2\%) & 82 (2.3\%) & 702 (13.8\%) & 115 (3.1\%) & 0 (0.0\%) & 0 (0.0\%) & \textbf{906 (4.8\%)} \\
\quad Test Interpretation & 4 (0.1\%) & 8 (0.2\%) & 593 (11.7\%) & 52 (1.4\%) & 0 (0.0\%) & 0 (0.0\%) & 657 (3.5\%) \\
\quad Test Selection & 3 (0.1\%) & 74 (2.1\%) & 109 (2.1\%) & 63 (1.7\%) & 0 (0.0\%) & 0 (0.0\%) & 249 (1.3\%) \\
\addlinespace[0.5ex]
\textbf{Prevention \& Screening} & 7 (0.2\%) & 117 (3.4\%) & 125 (2.5\%) & 264 (7.2\%) & 0 (0.0\%) & 0 (0.0\%) & \textbf{513 (2.7\%)} \\
\quad Vaccination & 3 (0.1\%) & 10 (0.3\%) & 94 (1.9\%) & 139 (3.8\%) & 0 (0.0\%) & 0 (0.0\%) & 246 (1.3\%) \\
\quad Lifestyle Prevention & 2 (0.1\%) & 89 (2.6\%) & 25 (0.5\%) & 66 (1.8\%) & 0 (0.0\%) & 0 (0.0\%) & 182 (1.0\%) \\
\quad Screening Schedule & 2 (0.1\%) & 18 (0.5\%) & 6 (0.1\%) & 59 (1.6\%) & 0 (0.0\%) & 0 (0.0\%) & 85 (0.5\%) \\
\addlinespace[0.5ex]
\textbf{Healthcare Navigation} & 1 (0.0\%) & 37 (1.1\%) & 182 (3.6\%) & 173 (4.7\%) & 0 (0.0\%) & 0 (0.0\%) & \textbf{393 (2.1\%)} \\
\quad Navigation Referral & 0 (0.0\%) & 19 (0.5\%) & 117 (2.3\%) & 66 (1.8\%) & 0 (0.0\%) & 0 (0.0\%) & 202 (1.1\%) \\
\quad Other Admin & 1 (0.0\%) & 9 (0.3\%) & 40 (0.8\%) & 30 (0.8\%) & 0 (0.0\%) & 0 (0.0\%) & 80 (0.4\%) \\
\quad Draft Communication & 0 (0.0\%) & 4 (0.1\%) & 10 (0.2\%) & 33 (0.9\%) & 0 (0.0\%) & 0 (0.0\%) & 47 (0.3\%) \\
\quad Insurance Billing & 0 (0.0\%) & 5 (0.1\%) & 13 (0.3\%) & 21 (0.6\%) & 0 (0.0\%) & 0 (0.0\%) & 39 (0.2\%) \\
\quad Document Summary & 0 (0.0\%) & 0 (0.0\%) & 2 (0.0\%) & 23 (0.6\%) & 0 (0.0\%) & 0 (0.0\%) & 25 (0.1\%) \\
\addlinespace[0.5ex]
\textbf{Find Medical Research} & 0 (0.0\%) & 21 (0.6\%) & 8 (0.2\%) & 170 (4.6\%) & 0 (0.0\%) & 0 (0.0\%) & \textbf{199 (1.1\%)} \\
\quad Study Evidence & 0 (0.0\%) & 7 (0.2\%) & 6 (0.1\%) & 100 (2.7\%) & 0 (0.0\%) & 0 (0.0\%) & 113 (0.6\%) \\
\quad Medical Guidelines & 0 (0.0\%) & 3 (0.1\%) & 1 (0.0\%) & 66 (1.8\%) & 0 (0.0\%) & 0 (0.0\%) & 70 (0.4\%) \\
\quad Research Explanation & 0 (0.0\%) & 11 (0.3\%) & 1 (0.0\%) & 4 (0.1\%) & 0 (0.0\%) & 0 (0.0\%) & 16 (0.1\%) \\
\addlinespace[0.5ex]
\textbf{Non Health} & 50 (1.6\%) & 12 (0.3\%) & 18 (0.4\%) & 5 (0.1\%) & 0 (0.0\%) & 0 (0.0\%) & \textbf{85 (0.5\%)} \\
\quad Offtopic Nonhealth & 50 (1.6\%) & 12 (0.3\%) & 18 (0.4\%) & 5 (0.1\%) & 0 (0.0\%) & 0 (0.0\%) & 85 (0.5\%) \\

\end{longtable}
\end{landscape}
}

\vspace{0.5cm}
{\small \emph{Values are presented as n (\%) for each dataset and overall total. Sub-intents are indented under their corresponding top-level intents.}}

\vspace{0.5cm}
{\small \emph{Values are presented as n (\%) for each dataset and overall total. Sub-intents are indented under their corresponding top-level intents.}}

\clearpage

\afterpage{%
\clearpage
\begin{landscape}
\small
\begin{longtable}{p{3.5cm}p{2.8cm}p{2.8cm}p{2.8cm}p{2.8cm}p{2.8cm}p{2.8cm}p{2.2cm}}
\caption{Prevalence of Clinical Domains and Specific Medical Conditions Across Benchmarks} \label{tab:topics} \\
\toprule
\textbf{Topic Area/Sub-Topic} & \textbf{HealthSearchQA} & \textbf{MashQA Test} & \textbf{MedRedQA} & \textbf{HealthBench} & \textbf{GoogleFitbit} & \textbf{GoogleFitbit} & \textbf{Total} \\
 &  &  & \textbf{Test} & \textbf{Main} & \textbf{Sleep} & \textbf{Fitness} & \\
\addlinespace[0.5ex]
\midrule
\endfirsthead
\multicolumn{8}{c}{\tablename\ \thetable\ -- \emph{Continued from previous page}} \\
\toprule
\textbf{Topic Area/Sub-Topic} & \textbf{HealthSearchQA} & \textbf{MashQA Test} & \textbf{MedRedQA} & \textbf{HealthBench} & \textbf{GoogleFitbit} & \textbf{GoogleFitbit} & \textbf{Total} \\
 &  &  & \textbf{Test} & \textbf{Main} & \textbf{Sleep} & \textbf{Fitness} & \\
\addlinespace[0.5ex]
\midrule
\endhead
\midrule
\multicolumn{8}{r}{\emph{Continued on next page}} \\
\endfoot
\bottomrule
\endlastfoot

\textbf{Holistic Health \& Wellness} & 63 (2.0\%) & 156 (4.5\%) & 108 (2.1\%) & 197 (5.3\%) & 1521 (100.0\%) & 1750 (100.0\%) & \textbf{3795 (20.3\%)} \\
\quad Sleep \& Lifestyle & 43 (1.4\%) & 31 (0.9\%) & 46 (0.9\%) & 71 (1.9\%) & 1521 (100.0\%) & 1750 (100.0\%) & 3462 (18.5\%) \\
\quad Alternative/Eastern & 0 (0.0\%) & 28 (0.8\%) & 6 (0.1\%) & 20 (0.5\%) & 0 (0.0\%) & 0 (0.0\%) & 54 (0.3\%) \\
\quad Fitness \& Exercise & 0 (0.0\%) & 14 (0.4\%) & 6 (0.1\%) & 24 (0.7\%) & 0 (0.0\%) & 0 (0.0\%) & 44 (0.2\%) \\
\quad Mental Wellness & 3 (0.1\%) & 19 (0.5\%) & 4 (0.1\%) & 5 (0.1\%) & 0 (0.0\%) & 0 (0.0\%) & 31 (0.2\%) \\
\quad Cosmeceuticals \& Topicals & 1 (0.0\%) & 7 (0.2\%) & 3 (0.1\%) & 6 (0.2\%) & 0 (0.0\%) & 0 (0.0\%) & 17 (0.1\%) \\
\addlinespace[0.5ex]
\textbf{Skin \& Hair} & 366 (11.5\%) & 272 (7.8\%) & 761 (15.0\%) & 315 (8.5\%) & 0 (0.0\%) & 0 (0.0\%) & \textbf{1714 (9.2\%)} \\
\quad Wounds & 36 (1.1\%) & 38 (1.1\%) & 175 (3.4\%) & 104 (2.8\%) & 0 (0.0\%) & 0 (0.0\%) & 353 (1.9\%) \\
\quad Rash & 56 (1.8\%) & 28 (0.8\%) & 194 (3.8\%) & 46 (1.2\%) & 0 (0.0\%) & 0 (0.0\%) & 324 (1.7\%) \\
\quad Infections & 64 (2.0\%) & 18 (0.5\%) & 132 (2.6\%) & 34 (0.9\%) & 0 (0.0\%) & 0 (0.0\%) & 248 (1.3\%) \\
\quad Eczema & 20 (0.6\%) & 29 (0.8\%) & 26 (0.5\%) & 17 (0.5\%) & 0 (0.0\%) & 0 (0.0\%) & 92 (0.5\%) \\
\quad Bites \& Stings & 11 (0.3\%) & 13 (0.4\%) & 42 (0.8\%) & 18 (0.5\%) & 0 (0.0\%) & 0 (0.0\%) & 84 (0.4\%) \\
\quad Acne & 5 (0.2\%) & 15 (0.4\%) & 19 (0.4\%) & 22 (0.6\%) & 0 (0.0\%) & 0 (0.0\%) & 61 (0.3\%) \\
\quad Hair Loss & 5 (0.2\%) & 1 (0.0\%) & 16 (0.3\%) & 21 (0.6\%) & 0 (0.0\%) & 0 (0.0\%) & 43 (0.2\%) \\
\addlinespace[0.5ex]
\textbf{Brain \& Nerves} & 366 (11.5\%) & 283 (8.1\%) & 454 (8.9\%) & 344 (9.3\%) & 0 (0.0\%) & 0 (0.0\%) & \textbf{1447 (7.7\%)} \\
\quad Headache/Migraine & 14 (0.4\%) & 100 (2.9\%) & 76 (1.5\%) & 88 (2.4\%) & 0 (0.0\%) & 0 (0.0\%) & 278 (1.5\%) \\
\quad Neuropathy & 43 (1.4\%) & 30 (0.9\%) & 63 (1.2\%) & 20 (0.5\%) & 0 (0.0\%) & 0 (0.0\%) & 156 (0.8\%) \\
\quad Cognitive Changes & 39 (1.2\%) & 21 (0.6\%) & 63 (1.2\%) & 31 (0.8\%) & 0 (0.0\%) & 0 (0.0\%) & 154 (0.8\%) \\
\quad Dizziness/Vertigo & 23 (0.7\%) & 8 (0.2\%) & 46 (0.9\%) & 59 (1.6\%) & 0 (0.0\%) & 0 (0.0\%) & 136 (0.7\%) \\
\quad Stroke/TIA & 23 (0.7\%) & 1 (0.0\%) & 42 (0.8\%) & 38 (1.0\%) & 0 (0.0\%) & 0 (0.0\%) & 104 (0.6\%) \\
\quad Seizure & 12 (0.4\%) & 7 (0.2\%) & 46 (0.9\%) & 14 (0.4\%) & 0 (0.0\%) & 0 (0.0\%) & 79 (0.4\%) \\
\addlinespace[0.5ex]
\textbf{Digestive \& Nutrition} & 252 (7.9\%) & 272 (7.8\%) & 521 (10.3\%) & 339 (9.2\%) & 0 (0.0\%) & 0 (0.0\%) & \textbf{1384 (7.4\%)} \\
\quad Liver Disease & 33 (1.0\%) & 32 (0.9\%) & 75 (1.5\%) & 24 (0.7\%) & 0 (0.0\%) & 0 (0.0\%) & 164 (0.9\%) \\
\quad Abdominal Pain & 12 (0.4\%) & 10 (0.3\%) & 69 (1.4\%) & 42 (1.1\%) & 0 (0.0\%) & 0 (0.0\%) & 133 (0.7\%) \\
\quad Reflux/Heartburn & 19 (0.6\%) & 6 (0.2\%) & 38 (0.7\%) & 33 (0.9\%) & 0 (0.0\%) & 0 (0.0\%) & 96 (0.5\%) \\
\quad Nausea \& Vomiting & 16 (0.5\%) & 2 (0.1\%) & 47 (0.9\%) & 16 (0.4\%) & 0 (0.0\%) & 0 (0.0\%) & 81 (0.4\%) \\
\quad Diarrhea & 7 (0.2\%) & 7 (0.2\%) & 37 (0.7\%) & 22 (0.6\%) & 0 (0.0\%) & 0 (0.0\%) & 73 (0.4\%) \\
\quad Inflammatory Bowel Disease (IBD) & 12 (0.4\%) & 39 (1.1\%) & 13 (0.3\%) & 9 (0.2\%) & 0 (0.0\%) & 0 (0.0\%) & 73 (0.4\%) \\
\quad Weight Management & 13 (0.4\%) & 4 (0.1\%) & 26 (0.5\%) & 27 (0.7\%) & 0 (0.0\%) & 0 (0.0\%) & 70 (0.4\%) \\
\quad Constipation & 4 (0.1\%) & 25 (0.7\%) & 28 (0.6\%) & 5 (0.1\%) & 0 (0.0\%) & 0 (0.0\%) & 62 (0.3\%) \\
\quad Gallbladder/Biliary & 8 (0.3\%) & 0 (0.0\%) & 18 (0.4\%) & 8 (0.2\%) & 0 (0.0\%) & 0 (0.0\%) & 34 (0.2\%) \\
\quad Infilammatory Bowel Disease (IBD) & 0 (0.0\%) & 0 (0.0\%) & 1 (0.0\%) & 0 (0.0\%) & 0 (0.0\%) & 0 (0.0\%) & 1 (0.0\%) \\
\addlinespace[0.5ex]
\textbf{Infections (General)} & 252 (7.9\%) & 109 (3.1\%) & 435 (8.6\%) & 425 (11.5\%) & 0 (0.0\%) & 0 (0.0\%) & \textbf{1221 (6.5\%)} \\
\quad COVID-19 & 2 (0.1\%) & 0 (0.0\%) & 139 (2.7\%) & 14 (0.4\%) & 0 (0.0\%) & 0 (0.0\%) & 155 (0.8\%) \\
\quad Fever (Unspecified) & 16 (0.5\%) & 7 (0.2\%) & 36 (0.7\%) & 66 (1.8\%) & 0 (0.0\%) & 0 (0.0\%) & 125 (0.7\%) \\
\quad Travel-Related & 20 (0.6\%) & 4 (0.1\%) & 4 (0.1\%) & 81 (2.2\%) & 0 (0.0\%) & 0 (0.0\%) & 109 (0.6\%) \\
\quad Antimicrobials/Antibiotics & 0 (0.0\%) & 7 (0.2\%) & 20 (0.4\%) & 27 (0.7\%) & 0 (0.0\%) & 0 (0.0\%) & 54 (0.3\%) \\
\quad Skin/Soft Tissue & 12 (0.4\%) & 5 (0.1\%) & 11 (0.2\%) & 5 (0.1\%) & 0 (0.0\%) & 0 (0.0\%) & 33 (0.2\%) \\
\quad Tuberculosis & 4 (0.1\%) & 1 (0.0\%) & 2 (0.0\%) & 11 (0.3\%) & 0 (0.0\%) & 0 (0.0\%) & 18 (0.1\%) \\
\quad Long COVID & 0 (0.0\%) & 0 (0.0\%) & 1 (0.0\%) & 3 (0.1\%) & 0 (0.0\%) & 0 (0.0\%) & 4 (0.0\%) \\
\addlinespace[0.5ex]
\textbf{Muscles, Bones \& Joints} & 274 (8.6\%) & 330 (9.5\%) & 340 (6.7\%) & 251 (6.8\%) & 0 (0.0\%) & 0 (0.0\%) & \textbf{1195 (6.4\%)} \\
\quad Arthritis & 32 (1.0\%) & 217 (6.2\%) & 21 (0.4\%) & 29 (0.8\%) & 0 (0.0\%) & 0 (0.0\%) & 299 (1.6\%) \\
\quad Back \& Neck Pain & 19 (0.6\%) & 30 (0.9\%) & 88 (1.7\%) & 80 (2.2\%) & 0 (0.0\%) & 0 (0.0\%) & 217 (1.2\%) \\
\quad Sprains \& Strains & 19 (0.6\%) & 9 (0.3\%) & 61 (1.2\%) & 39 (1.1\%) & 0 (0.0\%) & 0 (0.0\%) & 128 (0.7\%) \\
\quad Knee Problems & 12 (0.4\%) & 21 (0.6\%) & 19 (0.4\%) & 36 (1.0\%) & 0 (0.0\%) & 0 (0.0\%) & 88 (0.5\%) \\
\quad Shoulder Problems & 12 (0.4\%) & 11 (0.3\%) & 20 (0.4\%) & 13 (0.4\%) & 0 (0.0\%) & 0 (0.0\%) & 56 (0.3\%) \\
\quad Osteoporosis & 0 (0.0\%) & 1 (0.0\%) & 0 (0.0\%) & 0 (0.0\%) & 0 (0.0\%) & 0 (0.0\%) & 1 (0.0\%) \\
\addlinespace[0.5ex]
\textbf{Heart \& Circulation} & 190 (6.0\%) & 262 (7.5\%) & 428 (8.4\%) & 245 (6.6\%) & 0 (0.0\%) & 0 (0.0\%) & \textbf{1125 (6.0\%)} \\
\quad Chest Pain & 14 (0.4\%) & 11 (0.3\%) & 68 (1.3\%) & 57 (1.5\%) & 0 (0.0\%) & 0 (0.0\%) & 150 (0.8\%) \\
\quad Hypertension & 10 (0.3\%) & 50 (1.4\%) & 35 (0.7\%) & 44 (1.2\%) & 0 (0.0\%) & 0 (0.0\%) & 139 (0.7\%) \\
\quad Palpitations & 14 (0.4\%) & 0 (0.0\%) & 59 (1.2\%) & 26 (0.7\%) & 0 (0.0\%) & 0 (0.0\%) & 99 (0.5\%) \\
\quad Arrhythmia & 9 (0.3\%) & 26 (0.7\%) & 54 (1.1\%) & 8 (0.2\%) & 0 (0.0\%) & 0 (0.0\%) & 97 (0.5\%) \\
\quad Venous Thromboembolism & 21 (0.7\%) & 12 (0.3\%) & 39 (0.8\%) & 13 (0.4\%) & 0 (0.0\%) & 0 (0.0\%) & 85 (0.5\%) \\
\quad Heart Failure & 5 (0.2\%) & 49 (1.4\%) & 10 (0.2\%) & 3 (0.1\%) & 0 (0.0\%) & 0 (0.0\%) & 67 (0.4\%) \\
\quad Hyperlipidemia & 4 (0.1\%) & 26 (0.7\%) & 5 (0.1\%) & 25 (0.7\%) & 0 (0.0\%) & 0 (0.0\%) & 60 (0.3\%) \\
\quad Coronary Artery Disease & 0 (0.0\%) & 1 (0.0\%) & 0 (0.0\%) & 0 (0.0\%) & 0 (0.0\%) & 0 (0.0\%) & 1 (0.0\%) \\
\quad Coronary artery disease & 0 (0.0\%) & 0 (0.0\%) & 1 (0.0\%) & 0 (0.0\%) & 0 (0.0\%) & 0 (0.0\%) & 1 (0.0\%) \\
\addlinespace[0.5ex]
\textbf{Mental Health \& Psychiatry} & 226 (7.1\%) & 268 (7.7\%) & 259 (5.1\%) & 261 (7.1\%) & 0 (0.0\%) & 0 (0.0\%) & \textbf{1014 (5.4\%)} \\
\quad ADHD & 10 (0.3\%) & 121 (3.5\%) & 12 (0.2\%) & 31 (0.8\%) & 0 (0.0\%) & 0 (0.0\%) & 174 (0.9\%) \\
\quad Anxiety & 27 (0.9\%) & 20 (0.6\%) & 77 (1.5\%) & 49 (1.3\%) & 0 (0.0\%) & 0 (0.0\%) & 173 (0.9\%) \\
\quad Depression & 14 (0.4\%) & 25 (0.7\%) & 25 (0.5\%) & 47 (1.3\%) & 0 (0.0\%) & 0 (0.0\%) & 111 (0.6\%) \\
\quad Postpartum Mental Health & 6 (0.2\%) & 0 (0.0\%) & 1 (0.0\%) & 79 (2.1\%) & 0 (0.0\%) & 0 (0.0\%) & 86 (0.5\%) \\
\quad Substance Use & 0 (0.0\%) & 12 (0.3\%) & 30 (0.6\%) & 3 (0.1\%) & 0 (0.0\%) & 0 (0.0\%) & 45 (0.2\%) \\
\quad Bipolar Disorder & 4 (0.1\%) & 25 (0.7\%) & 10 (0.2\%) & 4 (0.1\%) & 0 (0.0\%) & 0 (0.0\%) & 43 (0.2\%) \\
\quad Eating Disorders & 18 (0.6\%) & 12 (0.3\%) & 8 (0.2\%) & 3 (0.1\%) & 0 (0.0\%) & 0 (0.0\%) & 41 (0.2\%) \\
\quad Suicidal Ideation (Crisis) & 0 (0.0\%) & 2 (0.1\%) & 26 (0.5\%) & 12 (0.3\%) & 0 (0.0\%) & 0 (0.0\%) & 40 (0.2\%) \\
\quad OCD & 19 (0.6\%) & 5 (0.1\%) & 10 (0.2\%) & 4 (0.1\%) & 0 (0.0\%) & 0 (0.0\%) & 38 (0.2\%) \\
\quad Schizophrenia & 5 (0.2\%) & 11 (0.3\%) & 7 (0.1\%) & 1 (0.0\%) & 0 (0.0\%) & 0 (0.0\%) & 24 (0.1\%) \\
\quad PTSD & 7 (0.2\%) & 1 (0.0\%) & 5 (0.1\%) & 4 (0.1\%) & 0 (0.0\%) & 0 (0.0\%) & 17 (0.1\%) \\
\addlinespace[0.5ex]
\textbf{Cancer} & 187 (5.9\%) & 524 (15.0\%) & 136 (2.7\%) & 85 (2.3\%) & 0 (0.0\%) & 0 (0.0\%) & \textbf{932 (5.0\%)} \\
\quad Breast & 20 (0.6\%) & 74 (2.1\%) & 15 (0.3\%) & 25 (0.7\%) & 0 (0.0\%) & 0 (0.0\%) & 134 (0.7\%) \\
\quad Skin & 9 (0.3\%) & 56 (1.6\%) & 26 (0.5\%) & 8 (0.2\%) & 0 (0.0\%) & 0 (0.0\%) & 99 (0.5\%) \\
\quad Lung & 6 (0.2\%) & 75 (2.1\%) & 6 (0.1\%) & 4 (0.1\%) & 0 (0.0\%) & 0 (0.0\%) & 91 (0.5\%) \\
\quad Colorectal & 9 (0.3\%) & 13 (0.4\%) & 5 (0.1\%) & 16 (0.4\%) & 0 (0.0\%) & 0 (0.0\%) & 43 (0.2\%) \\
\quad Prostate & 5 (0.2\%) & 15 (0.4\%) & 3 (0.1\%) & 6 (0.2\%) & 0 (0.0\%) & 0 (0.0\%) & 29 (0.2\%) \\
\addlinespace[0.5ex]
\textbf{Reproductive \& Sexual Health} & 170 (5.4\%) & 143 (4.1\%) & 368 (7.2\%) & 97 (2.6\%) & 0 (0.0\%) & 0 (0.0\%) & \textbf{778 (4.2\%)} \\
\quad STIs & 21 (0.7\%) & 24 (0.7\%) & 83 (1.6\%) & 12 (0.3\%) & 0 (0.0\%) & 0 (0.0\%) & 140 (0.7\%) \\
\quad Menstrual Disorders & 34 (1.1\%) & 11 (0.3\%) & 55 (1.1\%) & 13 (0.4\%) & 0 (0.0\%) & 0 (0.0\%) & 113 (0.6\%) \\
\quad Sexual Function & 30 (0.9\%) & 12 (0.3\%) & 56 (1.1\%) & 5 (0.1\%) & 0 (0.0\%) & 0 (0.0\%) & 103 (0.6\%) \\
\quad Contraception & 0 (0.0\%) & 4 (0.1\%) & 33 (0.6\%) & 44 (1.2\%) & 0 (0.0\%) & 0 (0.0\%) & 81 (0.4\%) \\
\quad Perimenopause/Menopause & 14 (0.4\%) & 35 (1.0\%) & 1 (0.0\%) & 3 (0.1\%) & 0 (0.0\%) & 0 (0.0\%) & 53 (0.3\%) \\
\quad Fertility & 16 (0.5\%) & 5 (0.1\%) & 15 (0.3\%) & 11 (0.3\%) & 0 (0.0\%) & 0 (0.0\%) & 47 (0.3\%) \\
\addlinespace[0.5ex]
\textbf{Endocrine \& Metabolic} & 120 (3.8\%) & 209 (6.0\%) & 183 (3.6\%) & 163 (4.4\%) & 0 (0.0\%) & 0 (0.0\%) & \textbf{675 (3.6\%)} \\
\quad Diabetes & 23 (0.7\%) & 106 (3.0\%) & 36 (0.7\%) & 79 (2.1\%) & 0 (0.0\%) & 0 (0.0\%) & 244 (1.3\%) \\
\quad Thyroid & 15 (0.5\%) & 13 (0.4\%) & 32 (0.6\%) & 19 (0.5\%) & 0 (0.0\%) & 0 (0.0\%) & 79 (0.4\%) \\
\quad Osteopenia/Osteoporosis & 11 (0.3\%) & 25 (0.7\%) & 0 (0.0\%) & 10 (0.3\%) & 0 (0.0\%) & 0 (0.0\%) & 46 (0.2\%) \\
\quad Obesity & 4 (0.1\%) & 13 (0.4\%) & 9 (0.2\%) & 14 (0.4\%) & 0 (0.0\%) & 0 (0.0\%) & 40 (0.2\%) \\
\quad PCOS & 7 (0.2\%) & 1 (0.0\%) & 7 (0.1\%) & 2 (0.1\%) & 0 (0.0\%) & 0 (0.0\%) & 17 (0.1\%) \\
\quad Lipids & 0 (0.0\%) & 1 (0.0\%) & 0 (0.0\%) & 0 (0.0\%) & 0 (0.0\%) & 0 (0.0\%) & 1 (0.0\%) \\
\addlinespace[0.5ex]
\textbf{Lungs \& Breathing (Colds/Flu/COVID)} & 109 (3.4\%) & 82 (2.3\%) & 232 (4.6\%) & 182 (4.9\%) & 0 (0.0\%) & 0 (0.0\%) & \textbf{605 (3.2\%)} \\
\quad Shortness of Breath & 7 (0.2\%) & 5 (0.1\%) & 54 (1.1\%) & 59 (1.6\%) & 0 (0.0\%) & 0 (0.0\%) & 125 (0.7\%) \\
\quad Upper Respiratory Infection & 15 (0.5\%) & 30 (0.9\%) & 22 (0.4\%) & 32 (0.9\%) & 0 (0.0\%) & 0 (0.0\%) & 99 (0.5\%) \\
\quad Cough & 12 (0.4\%) & 8 (0.2\%) & 30 (0.6\%) & 38 (1.0\%) & 0 (0.0\%) & 0 (0.0\%) & 88 (0.5\%) \\
\quad Pneumonia & 8 (0.3\%) & 16 (0.5\%) & 18 (0.4\%) & 15 (0.4\%) & 0 (0.0\%) & 0 (0.0\%) & 57 (0.3\%) \\
\quad COVID-19 & 7 (0.2\%) & 0 (0.0\%) & 46 (0.9\%) & 1 (0.0\%) & 0 (0.0\%) & 0 (0.0\%) & 54 (0.3\%) \\
\quad Wheezing & 4 (0.1\%) & 0 (0.0\%) & 6 (0.1\%) & 3 (0.1\%) & 0 (0.0\%) & 0 (0.0\%) & 13 (0.1\%) \\
\addlinespace[0.5ex]
\textbf{Ears, Nose \& Throat} & 141 (4.4\%) & 37 (1.1\%) & 221 (4.3\%) & 149 (4.0\%) & 0 (0.0\%) & 0 (0.0\%) & \textbf{548 (2.9\%)} \\
\quad Otitis/Ear Pain & 28 (0.9\%) & 9 (0.3\%) & 27 (0.5\%) & 36 (1.0\%) & 0 (0.0\%) & 0 (0.0\%) & 100 (0.5\%) \\
\quad Sore Throat & 17 (0.5\%) & 10 (0.3\%) & 47 (0.9\%) & 21 (0.6\%) & 0 (0.0\%) & 0 (0.0\%) & 95 (0.5\%) \\
\quad Hearing Loss/Tinnitus & 20 (0.6\%) & 5 (0.1\%) & 20 (0.4\%) & 19 (0.5\%) & 0 (0.0\%) & 0 (0.0\%) & 64 (0.3\%) \\
\quad Nasal Congestion & 9 (0.3\%) & 5 (0.1\%) & 18 (0.4\%) & 8 (0.2\%) & 0 (0.0\%) & 0 (0.0\%) & 40 (0.2\%) \\
\quad Sinusitis & 5 (0.2\%) & 0 (0.0\%) & 13 (0.3\%) & 17 (0.5\%) & 0 (0.0\%) & 0 (0.0\%) & 35 (0.2\%) \\
\addlinespace[0.5ex]
\textbf{Eyes \& Vision} & 138 (4.3\%) & 195 (5.6\%) & 111 (2.2\%) & 65 (1.8\%) & 0 (0.0\%) & 0 (0.0\%) & \textbf{509 (2.7\%)} \\
\quad Vision Changes & 39 (1.2\%) & 15 (0.4\%) & 54 (1.1\%) & 20 (0.5\%) & 0 (0.0\%) & 0 (0.0\%) & 128 (0.7\%) \\
\quad Dry Eye & 10 (0.3\%) & 13 (0.4\%) & 4 (0.1\%) & 6 (0.2\%) & 0 (0.0\%) & 0 (0.0\%) & 33 (0.2\%) \\
\quad Conjunctivitis & 9 (0.3\%) & 2 (0.1\%) & 4 (0.1\%) & 9 (0.2\%) & 0 (0.0\%) & 0 (0.0\%) & 24 (0.1\%) \\
\quad Foreign Body/Irritation & 1 (0.0\%) & 5 (0.1\%) & 15 (0.3\%) & 2 (0.1\%) & 0 (0.0\%) & 0 (0.0\%) & 23 (0.1\%) \\
\addlinespace[0.5ex]
\textbf{Kidney \& Urinary} & 115 (3.6\%) & 47 (1.3\%) & 186 (3.7\%) & 88 (2.4\%) & 0 (0.0\%) & 0 (0.0\%) & \textbf{436 (2.3\%)} \\
\quad Urinary Tract Infection & 19 (0.6\%) & 0 (0.0\%) & 52 (1.0\%) & 20 (0.5\%) & 0 (0.0\%) & 0 (0.0\%) & 91 (0.5\%) \\
\quad Chronic Kidney Disease & 7 (0.2\%) & 11 (0.3\%) & 11 (0.2\%) & 26 (0.7\%) & 0 (0.0\%) & 0 (0.0\%) & 55 (0.3\%) \\
\quad Incontinence & 20 (0.6\%) & 6 (0.2\%) & 10 (0.2\%) & 6 (0.2\%) & 0 (0.0\%) & 0 (0.0\%) & 42 (0.2\%) \\
\quad Hematuria & 8 (0.3\%) & 0 (0.0\%) & 22 (0.4\%) & 5 (0.1\%) & 0 (0.0\%) & 0 (0.0\%) & 35 (0.2\%) \\
\quad Kidney Stones & 8 (0.3\%) & 3 (0.1\%) & 9 (0.2\%) & 7 (0.2\%) & 0 (0.0\%) & 0 (0.0\%) & 27 (0.1\%) \\
\quad Prostatitis/BPH & 8 (0.3\%) & 2 (0.1\%) & 7 (0.1\%) & 2 (0.1\%) & 0 (0.0\%) & 0 (0.0\%) & 19 (0.1\%) \\

\end{longtable}
\end{landscape}
}

\vspace{0.5cm}
{\small \emph{Values are presented as n (\%) for each dataset and overall total. Sub-topics are indented under their corresponding topic areas.}}

\vspace{0.5cm}
{\small \emph{Values are presented as n (\%) for each dataset and overall total. Sub-topics are indented under their corresponding topic areas.}}

\clearpage

\afterpage{%
\clearpage
\begin{landscape}
\footnotesize
\begin{longtable}{p{3.5cm}p{1.2cm}p{1.2cm}p{2.2cm}p{2.2cm}p{2.2cm}p{2.2cm}p{2.2cm}}
\caption{Characteristics of Queries Mentioning Key Health Conditions (N=18,707 Consumer Queries)} \label{tab:key_conditions} \\
\toprule
\textbf{Condition} & \textbf{N} & \textbf{\% of All} & \textbf{Single-turn} & \textbf{Mod-High Risk} & \textbf{Generation 1} & \textbf{Generation 2} & \textbf{Generation 3} \\
\midrule
\endfirsthead
\multicolumn{8}{c}{\tablename\ \thetable\ -- \emph{Continued from previous page}} \\
\toprule
\textbf{Condition} & \textbf{N} & \textbf{\% of All} & \textbf{Single-turn} & \textbf{Mod-High Risk} & \textbf{Generation 1} & \textbf{Generation 2} & \textbf{Generation 3} \\
\midrule
\endhead
\midrule
\multicolumn{8}{r}{\emph{Continued on next page}} \\
\endfoot
\bottomrule
\endlastfoot

\textbf{Mental Health \& Behavioral} & \textbf{1798} & \textbf{9.61\%} & \textbf{1708 (95.0\%)} & \textbf{1163 (64.7\%)} & \textbf{279 (15.5\%)} & \textbf{1306 (72.6\%)} & \textbf{213 (11.8\%)} \\
\quad Anxiety & 685 & 3.66\% & 645 (94.2\%) & 482 (70.4\%) & 33 (4.8\%) & 566 (82.6\%) & 86 (12.6\%) \\
\quad Depression & 377 & 2.02\% & 358 (95.0\%) & 266 (70.6\%) & 33 (8.8\%) & 293 (77.7\%) & 51 (13.5\%) \\
\quad ADHD & 266 & 1.42\% & 251 (94.4\%) & 87 (32.7\%) & 128 (48.1\%) & 102 (38.3\%) & 36 (13.5\%) \\
\quad Bipolar Disorder & 85 & 0.45\% & 83 (97.6\%) & 41 (48.2\%) & 30 (35.3\%) & 50 (58.8\%) & 5 (5.9\%) \\
\quad Suicidal Ideation & 73 & 0.39\% & 69 (94.5\%) & 71 (97.3\%) & 2 (2.7\%) & 62 (84.9\%) & 9 (12.3\%) \\
\quad Eating Disorder & 63 & 0.34\% & 62 (98.4\%) & 34 (54.0\%) & 26 (41.3\%) & 34 (54.0\%) & 3 (4.8\%) \\
\quad Substance Use Disorder & 62 & 0.33\% & 61 (98.4\%) & 50 (80.6\%) & 4 (6.5\%) & 54 (87.1\%) & 4 (6.5\%) \\
\quad PTSD & 56 & 0.30\% & 54 (96.4\%) & 41 (73.2\%) & 4 (7.1\%) & 47 (83.9\%) & 5 (8.9\%) \\
\quad OCD & 52 & 0.28\% & 49 (94.2\%) & 32 (61.5\%) & 11 (21.2\%) & 38 (73.1\%) & 3 (5.8\%) \\
\quad Autism Spectrum Disorder & 43 & 0.23\% & 41 (95.3\%) & 23 (53.5\%) & 8 (18.6\%) & 29 (67.4\%) & 6 (14.0\%) \\
\quad Self Harm & 36 & 0.19\% & 35 (97.2\%) & 36 (100.0\%) & 0 (0.0\%) & 31 (86.1\%) & 5 (13.9\%) \\
\addlinespace[0.5ex]
\textbf{Endocrine \& Metabolic} & \textbf{741} & \textbf{3.96\%} & \textbf{677 (91.4\%)} & \textbf{347 (46.8\%)} & \textbf{183 (24.7\%)} & \textbf{422 (57.0\%)} & \textbf{136 (18.4\%)} \\
\quad Diabetes & 350 & 1.87\% & 310 (88.6\%) & 126 (36.0\%) & 138 (39.4\%) & 124 (35.4\%) & 88 (25.1\%) \\
\quad Obesity & 140 & 0.75\% & 136 (97.1\%) & 89 (63.6\%) & 11 (7.9\%) & 119 (85.0\%) & 10 (7.1\%) \\
\quad Thyroid Disease & 122 & 0.65\% & 111 (91.0\%) & 66 (54.1\%) & 20 (16.4\%) & 85 (69.7\%) & 17 (13.9\%) \\
\quad PCOS & 66 & 0.35\% & 65 (98.5\%) & 40 (60.6\%) & 7 (10.6\%) & 56 (84.8\%) & 3 (4.5\%) \\
\quad Hyperlipidemia & 63 & 0.34\% & 55 (87.3\%) & 26 (41.3\%) & 7 (11.1\%) & 38 (60.3\%) & 18 (28.6\%) \\
\addlinespace[0.5ex]
\textbf{Infectious Disease} & \textbf{509} & \textbf{2.72\%} & \textbf{479 (94.1\%)} & \textbf{255 (50.1\%)} & \textbf{58 (11.4\%)} & \textbf{374 (73.5\%)} & \textbf{77 (15.1\%)} \\
\quad COVID-19 & 365 & 1.95\% & 352 (96.4\%) & 198 (54.2\%) & 10 (2.7\%) & 329 (90.1\%) & 26 (7.1\%) \\
\quad HIV & 105 & 0.56\% & 98 (93.3\%) & 42 (40.0\%) & 42 (40.0\%) & 41 (39.0\%) & 22 (21.0\%) \\
\quad Malaria & 39 & 0.21\% & 29 (74.4\%) & 15 (38.5\%) & 6 (15.4\%) & 4 (10.3\%) & 29 (74.4\%) \\
\addlinespace[0.5ex]
\textbf{Maternal \& Reproductive} & \textbf{506} & \textbf{2.70\%} & \textbf{360 (71.1\%)} & \textbf{258 (51.0\%)} & \textbf{44 (8.7\%)} & \textbf{120 (23.7\%)} & \textbf{342 (67.6\%)} \\
\quad Postpartum & 241 & 1.29\% & 159 (66.0\%) & 127 (52.7\%) & 10 (4.1\%) & 37 (15.4\%) & 194 (80.5\%) \\
\quad Pregnant & 193 & 1.03\% & 158 (81.9\%) & 84 (43.5\%) & 31 (16.1\%) & 81 (42.0\%) & 81 (42.0\%) \\
\quad Postpartum Depression & 72 & 0.38\% & 43 (59.7\%) & 47 (65.3\%) & 3 (4.2\%) & 2 (2.8\%) & 67 (93.1\%) \\
\addlinespace[0.5ex]
\textbf{Cardiovascular} & \textbf{379} & \textbf{2.03\%} & \textbf{349 (92.1\%)} & \textbf{164 (43.3\%)} & \textbf{144 (38.0\%)} & \textbf{179 (47.2\%)} & \textbf{56 (14.8\%)} \\
\quad Hypertension & 229 & 1.22\% & 205 (89.5\%) & 113 (49.3\%) & 54 (23.6\%) & 130 (56.8\%) & 45 (19.7\%) \\
\quad Heart Failure & 68 & 0.36\% & 66 (97.1\%) & 18 (26.5\%) & 49 (72.1\%) & 15 (22.1\%) & 4 (5.9\%) \\
\quad Atrial Fibrillation & 51 & 0.27\% & 48 (94.1\%) & 18 (35.3\%) & 31 (60.8\%) & 16 (31.4\%) & 4 (7.8\%) \\
\quad Coronary Artery Disease & 31 & 0.17\% & 30 (96.8\%) & 15 (48.4\%) & 10 (32.3\%) & 18 (58.1\%) & 3 (9.7\%) \\
\addlinespace[0.5ex]
\textbf{Oncology} & \textbf{315} & \textbf{1.68\%} & \textbf{296 (94.0\%)} & \textbf{82 (26.0\%)} & \textbf{189 (60.0\%)} & \textbf{98 (31.1\%)} & \textbf{28 (8.9\%)} \\
\quad Breast Cancer & 103 & 0.55\% & 97 (94.2\%) & 20 (19.4\%) & 69 (67.0\%) & 24 (23.3\%) & 10 (9.7\%) \\
\quad Skin Cancer & 103 & 0.55\% & 100 (97.1\%) & 28 (27.2\%) & 61 (59.2\%) & 37 (35.9\%) & 5 (4.9\%) \\
\quad Colorectal Cancer & 42 & 0.22\% & 37 (88.1\%) & 17 (40.5\%) & 13 (31.0\%) & 21 (50.0\%) & 8 (19.0\%) \\
\quad Prostate Cancer & 34 & 0.18\% & 31 (91.2\%) & 10 (29.4\%) & 21 (61.8\%) & 10 (29.4\%) & 3 (8.8\%) \\
\quad Cervical Cancer & 33 & 0.18\% & 31 (93.9\%) & 7 (21.2\%) & 25 (75.8\%) & 6 (18.2\%) & 2 (6.1\%) \\
\addlinespace[0.5ex]
\textbf{Respiratory} & \textbf{282} & \textbf{1.51\%} & \textbf{256 (90.8\%)} & \textbf{155 (55.0\%)} & \textbf{69 (24.5\%)} & \textbf{158 (56.0\%)} & \textbf{55 (19.5\%)} \\
\quad Asthma & 212 & 1.13\% & 198 (93.4\%) & 123 (58.0\%) & 42 (19.8\%) & 136 (64.2\%) & 34 (16.0\%) \\
\quad COPD & 26 & 0.14\% & 23 (88.5\%) & 17 (65.4\%) & 4 (15.4\%) & 16 (61.5\%) & 6 (23.1\%) \\
\quad Tuberculosis & 25 & 0.13\% & 17 (68.0\%) & 14 (56.0\%) & 7 (28.0\%) & 4 (16.0\%) & 14 (56.0\%) \\
\quad Lung Cancer & 19 & 0.10\% & 18 (94.7\%) & 1 (5.3\%) & 16 (84.2\%) & 2 (10.5\%) & 1 (5.3\%) \\
\addlinespace[0.5ex]
\textbf{Renal \& Urologic} & \textbf{47} & \textbf{0.25\%} & \textbf{34 (72.3\%)} & \textbf{28 (59.6\%)} & \textbf{7 (14.9\%)} & \textbf{17 (36.2\%)} & \textbf{23 (48.9\%)} \\
\quad Chronic Kidney Disease & 47 & 0.25\% & 34 (72.3\%) & 28 (59.6\%) & 7 (14.9\%) & 17 (36.2\%) & 23 (48.9\%) \\
\addlinespace[0.5ex]
\textbf{Other} & \textbf{10} & \textbf{0.05\%} & \textbf{9 (90.0\%)} & \textbf{5 (50.0\%)} & \textbf{0 (0.0\%)} & \textbf{7 (70.0\%)} & \textbf{3 (30.0\%)} \\
\quad Long Covid & 10 & 0.05\% & 9 (90.0\%) & 5 (50.0\%) & 0 (0.0\%) & 7 (70.0\%) & 3 (30.0\%) \\

\end{longtable}
\end{landscape}
}

\vspace{0.5cm}
{\small \emph{Values are presented as n (\%) for each characteristic. Generation 1 = HealthSearchQA + MashQA; Generation 2 = MedRedQA; Generation 3 = HealthBench + GoogleFitbit. Conditions with fewer than 10 queries (1 condition: schizophrenia) are excluded from this table.}}

\vspace{0.5cm}
{\small \emph{Values are presented as n (\%) for each characteristic. Key conditions represent explicit mentions identified by the LLM tagger. Generation 1 = HealthSearchQA + MashQA; Generation 2 = MedRedQA; Generation 3 = HealthBench + GoogleFitbit. Conditions with fewer than 10 queries are excluded.}}

\clearpage
\subsection{Per-Dataset Query Profile Tables}
\label{sec:query-profile-tables}

This section presents detailed query profile tables for each individual dataset, showing the distribution of queries across five key dimensions: intent, topic, context richness, clinical complexity, and data integration. All profiles are based on consumer queries only.

\clearpage
\subsubsection{HealthSearchQA (N=3,173)}
{\small
\begin{longtable}{p{8cm}r}
\toprule
\textbf{Dimension/Category} & \textbf{Count (\%)} \\
\midrule
\endfirsthead

\multicolumn{2}{c}{\tablename\ \thetable\ -- \emph{Continued from previous page}} \\
\toprule
\textbf{Dimension/Category} & \textbf{Count (\%)} \\
\midrule
\endhead

\midrule
\multicolumn{2}{r}{\emph{Continued on next page}} \\
\endfoot

\bottomrule
\endlastfoot

\multicolumn{2}{l}{\textbf{\large 1. Intent Distribution}} \\
\addlinespace[0.5ex]
\textbf{Education} & 2,645 (83.4\%) \\
\quad Education Explainer & 2,635 (99.6\%) \\
\quad Basic Science & 10 (0.4\%) \\
\addlinespace[0.5ex]
\textbf{Symptom Check} & 380 (12.0\%) \\
\quad Management Plan & 235 (61.8\%) \\
\quad Differential Diagnosis & 107 (28.2\%) \\
\quad Triage Disposition & 38 (10.0\%) \\
\addlinespace[0.5ex]
\textbf{Non-Health} & 50 (1.6\%) \\
\quad Offtopic Nonhealth & 50 (100.0\%) \\
\addlinespace[0.5ex]
\textbf{General Health Advice} & 47 (1.5\%) \\
\quad Nutrition/Diet & 19 (40.4\%) \\
\quad Cosmeceuticals/Topicals & 10 (21.3\%) \\
\quad Sleep Hygiene & 8 (17.0\%) \\
\addlinespace[0.5ex]
\textbf{Condition Management} & 25 (0.8\%) \\
\quad Chronic Care Support & 21 (84.0\%) \\
\quad Acute Flare Management & 4 (16.0\%) \\
\addlinespace[0.5ex]
\textbf{Other Intents} (Aggregates 4 items) & 26 (0.8\%) \\
\addlinespace
\multicolumn{2}{l}{\textbf{\large 2. Topic Distribution (Top 5)}} \\
\addlinespace[0.5ex]
\textbf{Brain \& Nerves} & 366 (11.5\%) \\
\quad Other/Unspecified & 212 (57.9\%) \\
\quad Neuropathy & 43 (11.8\%) \\
\quad Cognitive Changes & 39 (10.7\%) \\
\addlinespace[0.5ex]
\textbf{Skin \& Hair} & 366 (11.5\%) \\
\quad Other/Unspecified & 169 (46.2\%) \\
\quad Infections & 64 (17.5\%) \\
\quad Rash & 56 (15.3\%) \\
\addlinespace[0.5ex]
\textbf{Muscles, Bones \& Joints} & 274 (8.6\%) \\
\quad Other/Unspecified & 180 (65.7\%) \\
\quad Arthritis & 32 (11.7\%) \\
\quad Sprains \& Strains & 19 (6.9\%) \\
\addlinespace[0.5ex]
\textbf{Digestive \& Nutrition} & 252 (7.9\%) \\
\quad Other/Unspecified & 128 (50.8\%) \\
\quad Liver Disease & 33 (13.1\%) \\
\quad Reflux/Heartburn & 19 (7.5\%) \\
\addlinespace[0.5ex]
\textbf{Infections (General)} & 252 (7.9\%) \\
\quad Other/Unspecified & 198 (78.6\%) \\
\quad Travel-Related & 20 (7.9\%) \\
\quad Fever (Unspecified) & 16 (6.3\%) \\
\addlinespace
\multicolumn{2}{l}{\textbf{\large 3. Context Richness}} \\
\addlinespace[0.5ex]
\textbf{Conversation Structure} & \\
\quad Single Turn & 3,173 (100.0\%) \\
\addlinespace[0.5ex]
\textbf{Narrative Detail} & \\
\quad Short & 3,173 (100.0\%) \\
\addlinespace[0.5ex]
\textbf{Context Depth} & \\
\quad Low & 3,171 (99.9\%) \\
\quad High & 2 (0.1\%) \\
\addlinespace
\multicolumn{2}{l}{\textbf{\large 4. Clinical Complexity}} \\
\addlinespace[0.5ex]
\textbf{Risk Level} & \\
\quad Low & 3,093 (97.5\%) \\
\quad Moderate & 70 (2.2\%) \\
\quad High & 10 (0.3\%) \\
\addlinespace[0.5ex]
\textbf{User Type} & \\
\quad Consumer & 3,173 (100.0\%) \\
\addlinespace[0.5ex]
\textbf{Population} & \\
\quad Adult (Unspecified) & 3,104 (97.8\%) \\
\quad Peds Unspecified & 40 (1.3\%) \\
\quad Pediatric (Under 5) & 26 (0.8\%) \\
\quad Adult 65Plus & 2 (0.1\%) \\
\quad Peds 5To17 & 1 (0.0\%) \\
\addlinespace[0.5ex]
\textbf{Language} & \\
\quad English & 3,173 (100.0\%) \\
\addlinespace[0.5ex]
\textbf{Language Complexity} & \\
\quad Lay & 3,069 (96.7\%) \\
\quad Technical & 104 (3.3\%) \\
\addlinespace[0.5ex]
\textbf{Query Subject} & \\
\quad General & 2,944 (92.8\%) \\
\quad Self & 218 (6.9\%) \\
\quad Child & 11 (0.3\%) \\
\addlinespace[0.5ex]
\textbf{Personal Health Query} & \\
\quad No & 2,955 (93.1\%) \\
\quad Yes & 218 (6.9\%) \\
\addlinespace
\multicolumn{2}{l}{\textbf{\large 5. Data Integration}} \\
\addlinespace[0.5ex]
\textbf{Objective Data Present} & \\
\quad Yes & 8 (0.2\%) \\
\quad No & 3,165 (99.8\%) \\
\addlinespace[0.5ex]
\textbf{Objective Data Types} & \\
\quad Diagnoses & 5 (0.2\%) \\
\quad Vitals (Basic) & 3 (0.1\%) \\

\end{longtable}

}

\clearpage
\subsubsection{MashQA Test (N=3,490)}
{\small
\begin{longtable}{p{8cm}r}
\toprule
\textbf{Dimension/Category} & \textbf{Count (\%)} \\
\midrule
\endfirsthead

\multicolumn{2}{c}{\tablename\ \thetable\ -- \emph{Continued from previous page}} \\
\toprule
\textbf{Dimension/Category} & \textbf{Count (\%)} \\
\midrule
\endhead

\midrule
\multicolumn{2}{r}{\emph{Continued on next page}} \\
\endfoot

\bottomrule
\endlastfoot

\multicolumn{2}{l}{\textbf{\large 1. Intent Distribution}} \\
\addlinespace[0.5ex]
\textbf{Education} & 2,389 (68.5\%) \\
\quad Education Explainer & 2,250 (94.2\%) \\
\quad Basic Science & 139 (5.8\%) \\
\addlinespace[0.5ex]
\textbf{Medication Information} & 332 (9.5\%) \\
\quad Side Effects & 157 (47.3\%) \\
\quad Selection & 112 (33.7\%) \\
\quad Dosing & 45 (13.6\%) \\
\addlinespace[0.5ex]
\textbf{General Health Advice} & 240 (6.9\%) \\
\quad Nutrition/Diet & 72 (30.0\%) \\
\quad Supplements/Nutraceuticals & 64 (26.7\%) \\
\quad Fitness/Exercise & 40 (16.7\%) \\
\addlinespace[0.5ex]
\textbf{Condition Management} & 157 (4.5\%) \\
\quad Chronic Care Support & 141 (89.8\%) \\
\quad Risk/Prognosis & 10 (6.4\%) \\
\quad Acute Flare Management & 6 (3.8\%) \\
\addlinespace[0.5ex]
\textbf{Prevention/Screening} & 117 (3.4\%) \\
\quad Lifestyle Prevention & 89 (76.1\%) \\
\quad Screening Schedule & 18 (15.4\%) \\
\quad Vaccination & 10 (8.6\%) \\
\addlinespace[0.5ex]
\textbf{Other Intents} (Aggregates 5 items) & 255 (7.3\%) \\
\addlinespace
\multicolumn{2}{l}{\textbf{\large 2. Topic Distribution (Top 5)}} \\
\addlinespace[0.5ex]
\textbf{Cancer} & 524 (15.0\%) \\
\quad Other/Unspecified & 291 (55.5\%) \\
\quad Lung & 75 (14.3\%) \\
\quad Breast & 74 (14.1\%) \\
\addlinespace[0.5ex]
\textbf{Muscles, Bones \& Joints} & 330 (9.5\%) \\
\quad Arthritis & 217 (65.8\%) \\
\quad Other/Unspecified & 41 (12.4\%) \\
\quad Back \& Neck Pain & 30 (9.1\%) \\
\addlinespace[0.5ex]
\textbf{Brain \& Nerves} & 283 (8.1\%) \\
\quad Other/Unspecified & 116 (41.0\%) \\
\quad Headache/Migraine & 100 (35.3\%) \\
\quad Neuropathy & 30 (10.6\%) \\
\addlinespace[0.5ex]
\textbf{Digestive \& Nutrition} & 272 (7.8\%) \\
\quad Other/Unspecified & 147 (54.0\%) \\
\quad Inflammatory Bowel Disease (IBD) & 39 (14.3\%) \\
\quad Liver Disease & 32 (11.8\%) \\
\addlinespace[0.5ex]
\textbf{Skin \& Hair} & 272 (7.8\%) \\
\quad Other/Unspecified & 130 (47.8\%) \\
\quad Wounds & 38 (14.0\%) \\
\quad Eczema & 29 (10.7\%) \\
\addlinespace
\multicolumn{2}{l}{\textbf{\large 3. Context Richness}} \\
\addlinespace[0.5ex]
\textbf{Conversation Structure} & \\
\quad Single Turn & 3,490 (100.0\%) \\
\addlinespace[0.5ex]
\textbf{Narrative Detail} & \\
\quad Short & 3,490 (100.0\%) \\
\addlinespace[0.5ex]
\textbf{Context Depth} & \\
\quad Low & 3,484 (99.8\%) \\
\quad High & 6 (0.2\%) \\
\addlinespace
\multicolumn{2}{l}{\textbf{\large 4. Clinical Complexity}} \\
\addlinespace[0.5ex]
\textbf{Risk Level} & \\
\quad Low & 3,441 (98.6\%) \\
\quad Moderate & 44 (1.3\%) \\
\quad High & 5 (0.1\%) \\
\addlinespace[0.5ex]
\textbf{User Type} & \\
\quad Consumer & 3,490 (100.0\%) \\
\addlinespace[0.5ex]
\textbf{Population} & \\
\quad Adult (Unspecified) & 3,297 (94.5\%) \\
\quad Peds Unspecified & 115 (3.3\%) \\
\quad Pediatric (Under 5) & 57 (1.6\%) \\
\quad Adult 65Plus & 13 (0.4\%) \\
\quad Peds 5To17 & 8 (0.2\%) \\
\addlinespace[0.5ex]
\textbf{Language} & \\
\quad English & 3,490 (100.0\%) \\
\addlinespace[0.5ex]
\textbf{Language Complexity} & \\
\quad Lay & 3,157 (90.5\%) \\
\quad Technical & 333 (9.5\%) \\
\addlinespace[0.5ex]
\textbf{Query Subject} & \\
\quad General & 3,215 (92.1\%) \\
\quad Self & 223 (6.4\%) \\
\quad Child & 52 (1.5\%) \\
\addlinespace[0.5ex]
\textbf{Personal Health Query} & \\
\quad No & 3,267 (93.6\%) \\
\quad Yes & 223 (6.4\%) \\
\addlinespace
\multicolumn{2}{l}{\textbf{\large 5. Data Integration}} \\
\addlinespace[0.5ex]
\textbf{Objective Data Present} & \\
\quad Yes & 133 (3.8\%) \\
\quad No & 3,357 (96.2\%) \\
\addlinespace[0.5ex]
\textbf{Objective Data Types} & \\
\quad Diagnoses & 110 (3.1\%) \\
\quad Medications & 21 (0.6\%) \\
\quad Procedures & 15 (0.4\%) \\
\quad Labs & 1 (0.0\%) \\

\end{longtable}

}

\clearpage
\subsubsection{MedRedQA Test (N=5,081)}
{\small
\begin{longtable}{p{8cm}r}
\toprule
\textbf{Dimension/Category} & \textbf{Count (\%)} \\
\midrule
\endfirsthead

\multicolumn{2}{c}{\tablename\ \thetable\ -- \emph{Continued from previous page}} \\
\toprule
\textbf{Dimension/Category} & \textbf{Count (\%)} \\
\midrule
\endhead

\midrule
\multicolumn{2}{r}{\emph{Continued on next page}} \\
\endfoot

\bottomrule
\endlastfoot

\multicolumn{2}{l}{\textbf{\large 1. Intent Distribution}} \\
\addlinespace[0.5ex]
\textbf{Symptom Check} & 2,656 (52.3\%) \\
\quad Differential Diagnosis & 1,235 (46.5\%) \\
\quad Management Plan & 774 (29.1\%) \\
\quad Triage Disposition & 647 (24.4\%) \\
\addlinespace[0.5ex]
\textbf{Tests and Results} & 702 (13.8\%) \\
\quad Test Interpretation & 593 (84.5\%) \\
\quad Test Selection & 109 (15.5\%) \\
\addlinespace[0.5ex]
\textbf{Medication Information} & 608 (12.0\%) \\
\quad Side Effects & 240 (39.5\%) \\
\quad Selection & 134 (22.0\%) \\
\quad Interactions & 119 (19.6\%) \\
\addlinespace[0.5ex]
\textbf{Condition Management} & 558 (11.0\%) \\
\quad Risk/Prognosis & 212 (38.0\%) \\
\quad Chronic Care Support & 209 (37.5\%) \\
\quad Acute Flare Management & 137 (24.6\%) \\
\addlinespace[0.5ex]
\textbf{Administrative Meta} & 182 (3.6\%) \\
\quad Navigation Referral & 117 (64.3\%) \\
\quad Other Admin & 40 (22.0\%) \\
\quad Insurance Billing & 13 (7.1\%) \\
\addlinespace[0.5ex]
\textbf{Other Intents} (Aggregates 5 items) & 375 (7.4\%) \\
\addlinespace
\multicolumn{2}{l}{\textbf{\large 2. Topic Distribution (Top 5)}} \\
\addlinespace[0.5ex]
\textbf{Skin \& Hair} & 761 (15.0\%) \\
\quad Rash & 194 (25.5\%) \\
\quad Wounds & 175 (23.0\%) \\
\quad Other/Unspecified & 157 (20.6\%) \\
\addlinespace[0.5ex]
\textbf{Digestive \& Nutrition} & 521 (10.2\%) \\
\quad Other/Unspecified & 169 (32.4\%) \\
\quad Liver Disease & 75 (14.4\%) \\
\quad Abdominal Pain & 69 (13.2\%) \\
\addlinespace[0.5ex]
\textbf{Brain \& Nerves} & 454 (8.9\%) \\
\quad Other/Unspecified & 118 (26.0\%) \\
\quad Headache/Migraine & 76 (16.7\%) \\
\quad Neuropathy & 63 (13.9\%) \\
\addlinespace[0.5ex]
\textbf{Infections (General)} & 435 (8.6\%) \\
\quad Other/Unspecified & 222 (51.0\%) \\
\quad COVID-19 & 139 (31.9\%) \\
\quad Fever (Unspecified) & 36 (8.3\%) \\
\addlinespace[0.5ex]
\textbf{Heart \& Circulation} & 428 (8.4\%) \\
\quad Other/Unspecified & 157 (36.7\%) \\
\quad Chest Pain & 68 (15.9\%) \\
\quad Palpitations & 59 (13.8\%) \\
\addlinespace
\multicolumn{2}{l}{\textbf{\large 3. Context Richness}} \\
\addlinespace[0.5ex]
\textbf{Conversation Structure} & \\
\quad Single Turn & 5,072 (99.8\%) \\
\quad Multi Turn & 9 (0.2\%) \\
\addlinespace[0.5ex]
\textbf{Narrative Detail} & \\
\quad Detailed & 4,804 (94.5\%) \\
\quad Short & 277 (5.5\%) \\
\addlinespace[0.5ex]
\textbf{Context Depth} & \\
\quad High & 4,461 (87.8\%) \\
\quad Low & 620 (12.2\%) \\
\addlinespace
\multicolumn{2}{l}{\textbf{\large 4. Clinical Complexity}} \\
\addlinespace[0.5ex]
\textbf{Risk Level} & \\
\quad Moderate & 2,942 (57.9\%) \\
\quad Low & 1,639 (32.3\%) \\
\quad High & 500 (9.8\%) \\
\addlinespace[0.5ex]
\textbf{User Type} & \\
\quad Consumer & 5,081 (100.0\%) \\
\addlinespace[0.5ex]
\textbf{Population} & \\
\quad Adult (Unspecified) & 4,394 (86.5\%) \\
\quad Peds 5To17 & 468 (9.2\%) \\
\quad Adult 65Plus & 117 (2.3\%) \\
\quad Pediatric (Under 5) & 97 (1.9\%) \\
\quad Peds Unspecified & 5 (0.1\%) \\
\addlinespace[0.5ex]
\textbf{Language} & \\
\quad English & 5,081 (100.0\%) \\
\addlinespace[0.5ex]
\textbf{Language Complexity} & \\
\quad Lay & 4,461 (87.8\%) \\
\quad Technical & 620 (12.2\%) \\
\addlinespace[0.5ex]
\textbf{Query Subject} & \\
\quad Self & 4,327 (85.2\%) \\
\quad Parent & 186 (3.7\%) \\
\quad Child & 160 (3.1\%) \\
\quad Partner & 158 (3.1\%) \\
\quad Other Relative & 116 (2.3\%) \\
\quad General & 85 (1.7\%) \\
\quad Friend Acquaintance & 48 (0.9\%) \\
\quad Patient & 1 (0.0\%) \\
\addlinespace[0.5ex]
\textbf{Personal Health Query} & \\
\quad Yes & 4,328 (85.2\%) \\
\quad No & 753 (14.8\%) \\
\addlinespace
\multicolumn{2}{l}{\textbf{\large 5. Data Integration}} \\
\addlinespace[0.5ex]
\textbf{Objective Data Present} & \\
\quad Yes & 3,646 (71.8\%) \\
\quad No & 1,435 (28.2\%) \\
\addlinespace[0.5ex]
\textbf{Objective Data Types} & \\
\quad Diagnoses & 2,230 (43.9\%) \\
\quad Medications & 2,044 (40.2\%) \\
\quad Labs & 898 (17.7\%) \\
\quad Procedures & 879 (17.3\%) \\
\quad Imaging & 659 (13.0\%) \\
\quad Vitals (Basic) & 383 (7.5\%) \\
\quad Vitals (Wearable) & 35 (0.7\%) \\

\end{longtable}

}

\clearpage
\subsubsection{HealthBench Main (N=3,692)}
{\small
\begin{longtable}{p{8cm}r}
\toprule
\textbf{Dimension/Category} & \textbf{Count (\%)} \\
\midrule
\endfirsthead

\multicolumn{2}{c}{\tablename\ \thetable\ -- \emph{Continued from previous page}} \\
\toprule
\textbf{Dimension/Category} & \textbf{Count (\%)} \\
\midrule
\endhead

\midrule
\multicolumn{2}{r}{\emph{Continued on next page}} \\
\endfoot

\bottomrule
\endlastfoot

\multicolumn{2}{l}{\textbf{\large 1. Intent Distribution}} \\
\addlinespace[0.5ex]
\textbf{Symptom Check} & 1,384 (37.5\%) \\
\quad Management Plan & 681 (49.2\%) \\
\quad Triage Disposition & 392 (28.3\%) \\
\quad Differential Diagnosis & 311 (22.5\%) \\
\addlinespace[0.5ex]
\textbf{Medication Information} & 729 (19.8\%) \\
\quad Selection & 443 (60.8\%) \\
\quad Dosing & 170 (23.3\%) \\
\quad Side Effects & 72 (9.9\%) \\
\addlinespace[0.5ex]
\textbf{Condition Management} & 297 (8.0\%) \\
\quad Chronic Care Support & 227 (76.4\%) \\
\quad Risk/Prognosis & 44 (14.8\%) \\
\quad Acute Flare Management & 26 (8.8\%) \\
\addlinespace[0.5ex]
\textbf{Education} & 285 (7.7\%) \\
\quad Education Explainer & 280 (98.2\%) \\
\quad Basic Science & 5 (1.8\%) \\
\addlinespace[0.5ex]
\textbf{General Health Advice} & 270 (7.3\%) \\
\quad Nutrition/Diet & 115 (42.6\%) \\
\quad Supplements/Nutraceuticals & 60 (22.2\%) \\
\quad Fitness/Exercise & 40 (14.8\%) \\
\addlinespace[0.5ex]
\textbf{Other Intents} (Aggregates 5 items) & 727 (19.7\%) \\
\addlinespace
\multicolumn{2}{l}{\textbf{\large 2. Topic Distribution (Top 5)}} \\
\addlinespace[0.5ex]
\textbf{Infections (General)} & 425 (11.5\%) \\
\quad Other/Unspecified & 218 (51.3\%) \\
\quad Travel-Related & 81 (19.1\%) \\
\quad Fever (Unspecified) & 66 (15.5\%) \\
\addlinespace[0.5ex]
\textbf{Brain \& Nerves} & 344 (9.3\%) \\
\quad Other/Unspecified & 94 (27.3\%) \\
\quad Headache/Migraine & 88 (25.6\%) \\
\quad Dizziness/Vertigo & 59 (17.1\%) \\
\addlinespace[0.5ex]
\textbf{Digestive \& Nutrition} & 339 (9.2\%) \\
\quad Other/Unspecified & 153 (45.1\%) \\
\quad Abdominal Pain & 42 (12.4\%) \\
\quad Reflux/Heartburn & 33 (9.7\%) \\
\addlinespace[0.5ex]
\textbf{Skin \& Hair} & 315 (8.5\%) \\
\quad Wounds & 104 (33.0\%) \\
\quad Other/Unspecified & 53 (16.8\%) \\
\quad Rash & 46 (14.6\%) \\
\addlinespace[0.5ex]
\textbf{Mental Health \& Psychiatry} & 261 (7.1\%) \\
\quad Postpartum Mental Health & 79 (30.3\%) \\
\quad Anxiety & 49 (18.8\%) \\
\quad Depression & 47 (18.0\%) \\
\addlinespace
\multicolumn{2}{l}{\textbf{\large 3. Context Richness}} \\
\addlinespace[0.5ex]
\textbf{Conversation Structure} & \\
\quad Single Turn & 2,193 (59.4\%) \\
\quad Multi Turn & 1,499 (40.6\%) \\
\addlinespace[0.5ex]
\textbf{Narrative Detail} & \\
\quad Short & 3,237 (87.7\%) \\
\quad Detailed & 455 (12.3\%) \\
\addlinespace[0.5ex]
\textbf{Context Depth} & \\
\quad Low & 3,174 (86.0\%) \\
\quad High & 518 (14.0\%) \\
\addlinespace
\multicolumn{2}{l}{\textbf{\large 4. Clinical Complexity}} \\
\addlinespace[0.5ex]
\textbf{Risk Level} & \\
\quad Low & 2,161 (58.5\%) \\
\quad Moderate & 1,189 (32.2\%) \\
\quad High & 342 (9.3\%) \\
\addlinespace[0.5ex]
\textbf{User Type} & \\
\quad Consumer & 3,692 (100.0\%) \\
\addlinespace[0.5ex]
\textbf{Population} & \\
\quad Adult (Unspecified) & 3,249 (88.0\%) \\
\quad Pediatric (Under 5) & 159 (4.3\%) \\
\quad Peds Unspecified & 121 (3.3\%) \\
\quad Peds 5To17 & 93 (2.5\%) \\
\quad Adult 65Plus & 70 (1.9\%) \\
\addlinespace[0.5ex]
\textbf{Language} & \\
\quad English & 3,010 (81.5\%) \\
\quad Non-English & 682 (18.5\%) \\
\addlinespace[0.5ex]
\textbf{Language Complexity} & \\
\quad Lay & 3,470 (94.0\%) \\
\quad Technical & 222 (6.0\%) \\
\addlinespace[0.5ex]
\textbf{Query Subject} & \\
\quad Self & 2,247 (60.9\%) \\
\quad General & 870 (23.6\%) \\
\quad Child & 309 (8.4\%) \\
\quad Friend Acquaintance & 99 (2.7\%) \\
\quad Parent & 75 (2.0\%) \\
\quad Other Relative & 67 (1.8\%) \\
\quad Partner & 23 (0.6\%) \\
\quad Patient & 2 (0.1\%) \\
\addlinespace[0.5ex]
\textbf{Personal Health Query} & \\
\quad Yes & 2,247 (60.9\%) \\
\quad No & 1,445 (39.1\%) \\
\addlinespace
\multicolumn{2}{l}{\textbf{\large 5. Data Integration}} \\
\addlinespace[0.5ex]
\textbf{Objective Data Present} & \\
\quad Yes & 863 (23.4\%) \\
\quad No & 2,829 (76.6\%) \\
\addlinespace[0.5ex]
\textbf{Objective Data Types} & \\
\quad Diagnoses & 546 (14.8\%) \\
\quad Medications & 257 (7.0\%) \\
\quad Procedures & 99 (2.7\%) \\
\quad Labs & 82 (2.2\%) \\
\quad Vitals (Basic) & 66 (1.8\%) \\
\quad Imaging & 43 (1.2\%) \\
\quad Vitals (Wearable) & 1 (0.0\%) \\

\end{longtable}

}

\clearpage
\subsubsection{GoogleFitbit Sleep (N=1,521)}
{\small
\begin{longtable}{p{8cm}r}
\toprule
\textbf{Dimension/Category} & \textbf{Count (\%)} \\
\midrule
\endfirsthead

\multicolumn{2}{c}{\tablename\ \thetable\ -- \emph{Continued from previous page}} \\
\toprule
\textbf{Dimension/Category} & \textbf{Count (\%)} \\
\midrule
\endhead

\midrule
\multicolumn{2}{r}{\emph{Continued on next page}} \\
\endfoot

\bottomrule
\endlastfoot

\multicolumn{2}{l}{\textbf{\large 1. Intent Distribution}} \\
\addlinespace[0.5ex]
\textbf{General Health Advice} & 1,521 (100.0\%) \\
\quad Sleep Hygiene & 1,521 (100.0\%) \\
\addlinespace
\multicolumn{2}{l}{\textbf{\large 2. Topic Distribution (Top 5)}} \\
\addlinespace[0.5ex]
\textbf{Holistic Health \& Wellness} & 1,521 (100.0\%) \\
\quad Sleep \& Lifestyle & 1,521 (100.0\%) \\
\addlinespace
\multicolumn{2}{l}{\textbf{\large 3. Context Richness}} \\
\addlinespace[0.5ex]
\textbf{Conversation Structure} & \\
\quad Single Turn & 1,521 (100.0\%) \\
\addlinespace[0.5ex]
\textbf{Narrative Detail} & \\
\quad Detailed & 1,521 (100.0\%) \\
\addlinespace[0.5ex]
\textbf{Context Depth} & \\
\quad High & 1,521 (100.0\%) \\
\addlinespace
\multicolumn{2}{l}{\textbf{\large 4. Clinical Complexity}} \\
\addlinespace[0.5ex]
\textbf{Risk Level} & \\
\quad Low & 1,521 (100.0\%) \\
\addlinespace[0.5ex]
\textbf{User Type} & \\
\quad Consumer & 1,521 (100.0\%) \\
\addlinespace[0.5ex]
\textbf{Population} & \\
\quad Adult (Unspecified) & 1,092 (71.8\%) \\
\quad Adult 65Plus & 429 (28.2\%) \\
\addlinespace[0.5ex]
\textbf{Language} & \\
\quad English & 1,521 (100.0\%) \\
\addlinespace[0.5ex]
\textbf{Language Complexity} & \\
\quad Lay & 1,521 (100.0\%) \\
\addlinespace[0.5ex]
\textbf{Query Subject} & \\
\quad Self & 1,521 (100.0\%) \\
\addlinespace[0.5ex]
\textbf{Personal Health Query} & \\
\quad Yes & 1,521 (100.0\%) \\
\addlinespace
\multicolumn{2}{l}{\textbf{\large 5. Data Integration}} \\
\addlinespace[0.5ex]
\textbf{Objective Data Present} & \\
\quad Yes & 1,521 (100.0\%) \\
\addlinespace[0.5ex]
\textbf{Objective Data Types} & \\
\quad Vitals (Wearable) & 1,521 (100.0\%) \\

\end{longtable}

}

\clearpage
\subsubsection{GoogleFitbit Fitness (N=1,750)}
{\small
\begin{longtable}{p{8cm}r}
\toprule
\textbf{Dimension/Category} & \textbf{Count (\%)} \\
\midrule
\endfirsthead

\multicolumn{2}{c}{\tablename\ \thetable\ -- \emph{Continued from previous page}} \\
\toprule
\textbf{Dimension/Category} & \textbf{Count (\%)} \\
\midrule
\endhead

\midrule
\multicolumn{2}{r}{\emph{Continued on next page}} \\
\endfoot

\bottomrule
\endlastfoot

\multicolumn{2}{l}{\textbf{\large 1. Intent Distribution}} \\
\addlinespace[0.5ex]
\textbf{General Health Advice} & 1,750 (100.0\%) \\
\quad Fitness/Exercise & 1,750 (100.0\%) \\
\addlinespace
\multicolumn{2}{l}{\textbf{\large 2. Topic Distribution (Top 5)}} \\
\addlinespace[0.5ex]
\textbf{Holistic Health \& Wellness} & 1,750 (100.0\%) \\
\quad Sleep \& Lifestyle & 1,750 (100.0\%) \\
\addlinespace
\multicolumn{2}{l}{\textbf{\large 3. Context Richness}} \\
\addlinespace[0.5ex]
\textbf{Conversation Structure} & \\
\quad Single Turn & 1,750 (100.0\%) \\
\addlinespace[0.5ex]
\textbf{Narrative Detail} & \\
\quad Detailed & 1,750 (100.0\%) \\
\addlinespace[0.5ex]
\textbf{Context Depth} & \\
\quad High & 1,750 (100.0\%) \\
\addlinespace
\multicolumn{2}{l}{\textbf{\large 4. Clinical Complexity}} \\
\addlinespace[0.5ex]
\textbf{Risk Level} & \\
\quad Low & 1,750 (100.0\%) \\
\addlinespace[0.5ex]
\textbf{User Type} & \\
\quad Consumer & 1,750 (100.0\%) \\
\addlinespace[0.5ex]
\textbf{Population} & \\
\quad Adult (Unspecified) & 1,520 (86.9\%) \\
\quad Adult 65Plus & 230 (13.1\%) \\
\addlinespace[0.5ex]
\textbf{Language} & \\
\quad English & 1,750 (100.0\%) \\
\addlinespace[0.5ex]
\textbf{Language Complexity} & \\
\quad Lay & 1,750 (100.0\%) \\
\addlinespace[0.5ex]
\textbf{Query Subject} & \\
\quad Self & 1,750 (100.0\%) \\
\addlinespace[0.5ex]
\textbf{Personal Health Query} & \\
\quad Yes & 1,750 (100.0\%) \\
\addlinespace
\multicolumn{2}{l}{\textbf{\large 5. Data Integration}} \\
\addlinespace[0.5ex]
\textbf{Objective Data Present} & \\
\quad Yes & 1,750 (100.0\%) \\
\addlinespace[0.5ex]
\textbf{Objective Data Types} & \\
\quad Vitals (Wearable) & 1,750 (100.0\%) \\

\end{longtable}

}

\end{document}